\documentclass[times,twocolumn,final]{elsarticle}
\usepackage{etoolbox}
\AtBeginDocument{%
	\geometry{twoside, inner=43pt, outer=32pt}
}

\usepackage{cag}
\usepackage{framed,multirow}

\usepackage{amssymb}
\usepackage{amsmath}
\usepackage{bm}
\usepackage[percent]{overpic}
\usepackage[ruled]{algorithm2e}
\usepackage{booktabs} 
\usepackage{latexsym}

\usepackage{url}
\usepackage{xcolor}
\definecolor{newcolor}{rgb}{.8,.349,.1}

\usepackage{hyperref}
\usepackage{graphicx}

\newcommand{\orcidicon}[1]{\href{https://orcid.org/#1}{\includegraphics[height=1.6ex]{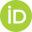}}}


\usepackage[switch,pagewise]{lineno} 
\usepackage{orcidlink}
\journal{Multimodal Conditional 3D Face Geometry Generation}
\newcommand{\figref}[1]{Fig.~\ref{#1}}
\newcommand{\tabref}[1]{Table~\ref{#1}}
\newcommand{\eqnref}[1]{Eq.~\ref{#1}}
\newcommand{\secref}[1]{Section~\ref{#1}}

\newcommand{\etal}{et al.}
\newcommand{\eg}{e.g.}

\definecolor{cocolor}{RGB}{210,10,210}

\begin{document}

\verso{Otto et al.}

\begin{frontmatter}

\title{Multimodal Conditional 3D Face Geometry Generation}%

\author[]{Christopher Otto\textsuperscript{a,b}\orcidlink{0000-0002-5625-593X}}
\author[]{Prashanth Chandran\textsuperscript{b}\orcidlink{0000-0001-6821-5815}}
\author[]{Sebastian Weiss\textsuperscript{b}\orcidlink{0000-0003-4399-3180}}
\author[]{Markus Gross\textsuperscript{a,b}\orcidlink{0009-0003-9324-779X}}
\author[]{Gaspard Zoss\textsuperscript{b}\orcidlink{0000-0002-0022-8203}}
\author[]{Derek Bradley\textsuperscript{b}\orcidlink{0000-0002-2055-9325}}

\address[1]{ETH Zürich, Switzerland}
\address[2]{DisneyResearch$|$Studios, Switzerland}

\accepted{24 July 2025}

\begin{abstract}
	We present a new method for multimodal conditional 3D face geometry generation that allows user-friendly control over the output identity and expression via a number of different conditioning signals.  Within a single model, we demonstrate 3D faces generated from artistic sketches, portrait photos, Canny edges, FLAME face model parameters, 2D face landmarks, or text prompts.  Our approach is based on a diffusion process that generates 3D geometry in a 2D parameterized UV domain.  Geometry generation passes each conditioning signal through a set of cross-attention layers (IP-Adapter), one set for each user-defined conditioning signal. The result is an easy-to-use 3D face generation tool that produces topology-consistent, high-quality geometry with fine-grain user control.
\end{abstract}


\end{frontmatter}


\section{Introduction}
\label{sec:introduction}

The creation of 3D facial geometry for digital human characters is a modeling task that usually requires tremendous artistic skill.  Digital sculpting with 3D modeling tools is a time-consuming and demanding process, especially when the target is as recognizable as a human face.  This complexity has prompted research into data-driven sculpting methods~\cite{sculpting} and other, more user-friendly, interactive interfaces~\cite{interactive_sketching}.

Several common morphable 3D face models (e.g. FLAME~\cite{flame}) simplify the facial modeling task by providing a shape subspace to operate in, as well as simple parameters to control the identity and expression geometry without the need for 3D modeling skills, but they are limited in expressiveness and offer only basic control knobs.

Recent methods can create high quality 3D geometry and textures from text prompts~\cite{dreamface,faceg2e,headartist,HeadSculpt,HeadEvolver} via optimization, leveraging large pre-trained text-to-image diffusion models~\cite{ldm}.  These methods allow layman users to create 3D faces through natural text descriptions.  While this is a powerful approach, it can still be difficult to achieve a particular output through text description~\cite{allaboutyoursketch}.  Some concepts like the specific curvature of a face or a unique facial expression are much easier to convey via sketches, edge contours or portrait photos than through text.

In the image domain, approaches like ControlNet~\cite{controlnet} or T2I-Adapter~\cite{t2iadapter} have demonstrated controllable image generation beyond text using sketches, images, or edge maps as conditioning signals. These methods provide users with much more fine-grained control over the generation process than text-based methods alone. Ye \etal~\cite{ipadapter} propose IP-Adapters to control Stable Diffusion~\cite{ldm} with image prompts by learning new cross-attention layers. However, image-based methods are not easy to extend to 3D facial geometry generation.

\begin{figure*}
	\centering
	\includegraphics[width=0.9\textwidth]{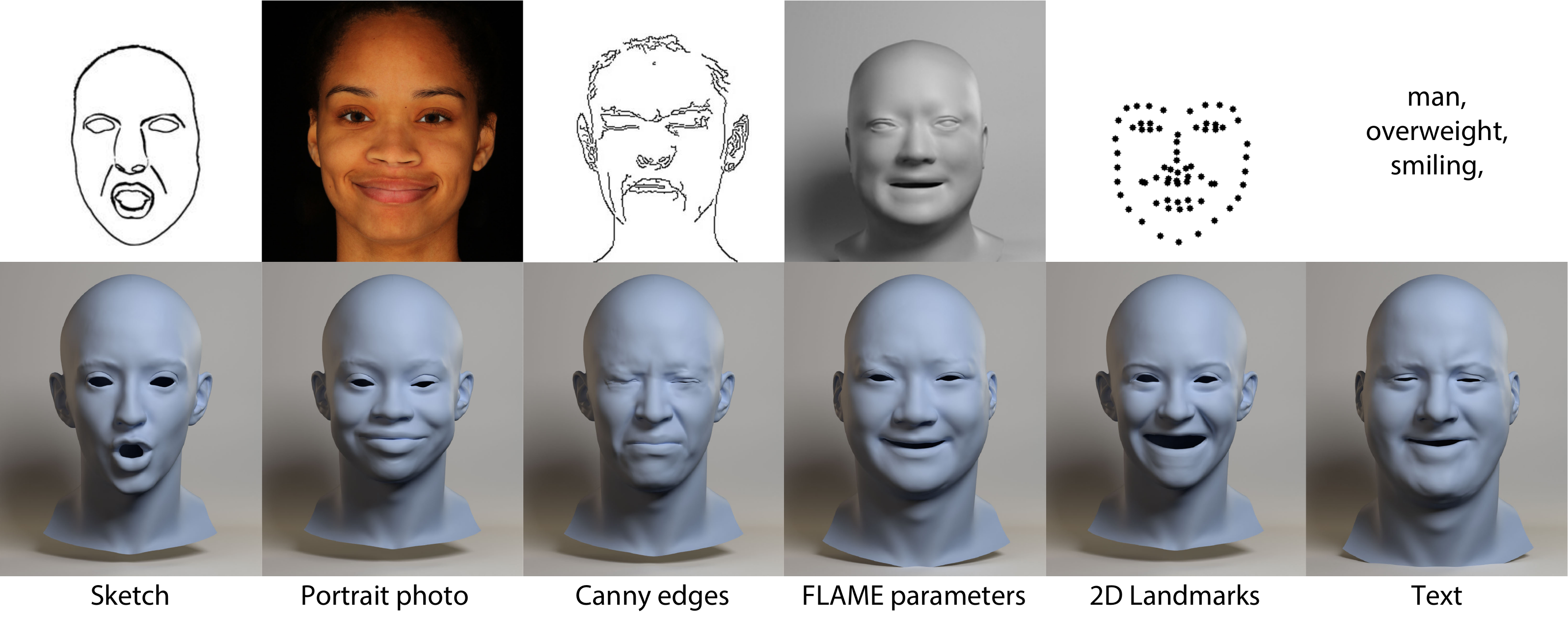}
	\caption{We propose a novel method for diffusion-based controllable 3D face geometry generation that allows for controlling the results via several conditioning modes: artistic sketches, portrait photos, Canny edges, FLAME parameters, 2D facial landmarks, and text.
	}
	\label{fig:teaser}
\end{figure*}

We present a flexible new method for 3D facial geometry generation that creates high quality faces from any one of various inputs, including sketches, 2D landmarks, Canny edges, FLAME face model parameters, portrait photos and text.  
Our approach is to train a conditional diffusion model on a high quality 3D facial dataset constructed from high resolution scans~\cite{sdfm} represented in the 2D UV domain. Our model is trained from scratch, without the need for a pre-trained foundation model like Stable Diffusion. To condition our model we train one set of cross-attention layers for each type of conditioning input, following IP-Adapter~\cite{ipadapter}. First, the diffusion model learns to inject FLAME parameters via the original UNet cross-attention layers. We then freeze the diffusion model while training additional sets of cross-attention layers (e.g. one for artist sketches, 2D landmarks, portrait photos, etc). Our FLAME-conditioned model allows us to re-interpret FLAME-parameterized faces in a generative sense - providing a space of high resolution stochastic variations on top of the traditional low resolution FLAME model.

Our method allows for fast and user-friendly creation of 3D digital character faces with expressions, generated with a consistent mesh topology, and controlled by the input mode preferred by the user, all within a single model.  We demonstrate several applications of our method including sketch-based 3D face modeling, geometry from 2D facial landmarks, Canny edges, or portrait photos, text-to-3D facial geometry, and finally, extending the FLAME model space by allowing stochastic diffusion sampling conditioned on the same semantic FLAME parameters (\figref{fig:teaser}).
In summary, we make the following contributions:
\begin{itemize}
	\item We present a new method for 3D face geometry generation from 6 different types of conditionings (prompts) within a single model.
	\item We propose a comprehensive solution for training such a method from scratch, with 3D geometry data augmentations and by representing 3D geometry as position maps to better fit existing diffusion pipelines.
	\item We show that our method supports face generation with expressions, sketch-based editing for 3D face design, stochastic variations of details conditioned on low resolution FLAME faces, generalization to in-the-wild data and dynamic face generation from videos.
\end{itemize}
\section{Related Work}
\label{sec:related_work}
In the following, we present relevant related work on 3D face geometry generation with diffusion models, as well as on injecting additional control modes into diffusion models.

\subsection{3D Face Geometry Generation}

Recent work uses diffusion models to control the generation of novel 3D face geometry.
ShapeFusion~\cite{shapefusion} generates face geometry by running the diffusion process directly on the mesh input vertices. It allows unconditional and conditional face geometry generation and supports various editing operations on a given mesh, based on selected vertices (anchor points). However, it does not support conditioning signals beyond vertices.
4DFM~\cite{4dfm} trains an unconditional diffusion model on a set of sparse 3D landmarks for facial expression generation. It can generate dynamic facial expression sequences based on 3D landmarks by retargeting the landmarks to a mesh after the diffusion process. While they support conditioning with different signals such as expression labels and text, they achieve control via classifier-guidance~\cite{classifier_guidance} which requires training additional classifiers on noisy data.

Other methods focus on 3D face or head avatars, which can generate 3D geometry and texture. Rodin~\cite{rodin} can generate triplane-based head geometry with text or image conditioning, but the resulting geometry is extracted with Marching Cubes~\cite{MarchingCubes} and thus not in the same topology across generations. HeadArtist~\cite{headartist}, HeadSculpt~\cite{HeadSculpt} and HumanNorm~\cite{humannorm} use Deep Marching Tetrahedra~\cite{dmtet}, text-prompts and a Score Distillation Loss (SDS)~\cite{sdsloss} to generate high-quality human heads. However, the topology of the extracted geometry differs across samples and generating a single geometry sample takes almost one hour on a single 3090 GPU.
DreamFace~\cite{dreamface}, FaceG2E~\cite{faceg2e} and Bergman \etal~\cite{art3dhead} propose 3D Morphable Model (3DMM)-based~\cite{3dmm} pipelines that generate topology-consistent 3D face geometry and textures from text. The geometry is created by optimizing 3DMM parameters using a SDS loss. During the optimization, the SDS loss uses the feedback from a pre-trained text-to-image latent diffusion model to update the 3DMM parameters given a geometry render. DreamFace~\cite{dreamface} and FaceG2E~\cite{faceg2e} focus on optimizing 3DMM identity parameters and have therefore difficulties in directly generating facial expressions from text prompts. However, DreamFace does support generating faces with expressions from image prompts and FaceG2E results can be imported into the CG pipeline where facial expressions can be added in a separate step after the text-based generation.
In general, SDS-based methods rely on the Stable Diffusion (SD)~\cite{ldm} prior which was pre-trained on billions of images~\cite{ldm_data} to guide their generations. SDS optimization is usually much slower in runtime compared to standard diffusion sampling, as it must backpropagate gradients from a diffusion model to a 3D model via a differentiable renderer for many optimization steps. Describe3D~\cite{describe3d} can generate 3D face geometry from text-prompts without diffusion, by mapping CLIP text embeddings to 3DMM shape parameters. However, it does not support different facial expressions.

In our work, we generate controllable 3D face geometry in a single common topology from several different conditioning modes. Our method natively supports facial expressions and does not rely on SDS optimization, classifier-guidance or the SD prior.

\subsection{Multimodal Conditional Image Generation}

Image generation with diffusion models can be controlled with conditioning modes (prompt types) that are different from text such as sketches~\cite{sketch_conditioning}, Canny edges~\cite{canny}, RGB images~\cite{ipadapter}, expression parameters~\cite{diffusionavatars} or face shape~\cite{diffusionrig,control3diff}. To control pre-trained diffusion models with new modes, ControlNet~\cite{controlnet} introduces a trainable copy of the diffusion model's UNet encoding blocks, which take the new conditioning as input. The output of the copied model is added to the skip-connections of the frozen pre-trained diffusion model. A separate trainable copy of the UNet encoding blocks (361M parameters) is created per conditioning mode.
T2I-Adapter~\cite{t2iadapter} aligns the internal knowledge of a pre-trained diffusion model with new control modes by proposing a small adapter network that achieves control similar to ControlNet, while requiring less parameters (77M).
IP-Adapter~\cite{ipadapter} injects each conditioning via separate cross-attention layers~\cite{attention} while requiring even less parameters (22M). It introduces new cross-attention layers whose outputs are added to those of the original UNet. Ye \etal~\cite{ipadapter} show that the diffusion model follows the added conditioning signal closely, when it comes through the newly trained cross-attention layers.
In general, adapters can add control with new conditioning modes even long after the training of the underlying base diffusion model is concluded. They can also add new conditioning modes for which only limited paired training data is available, because the underlying diffusion model is frozen, avoiding issues such as catastrophic forgetting~\cite{catastrophicforgetting,controlnet}.
While pre-trained text-to-image diffusion models understand the RGB image domain and can generate images conditioned on a large variety of input modalities, the domain gap to 3D face geometry is large. Therefore, many related works use the rather slow SDS optimization procedure to lift faces into 3D. To allow for faster inference-time sampling, we train our diffusion model from scratch using ground truth 3D face data and geometry data augmentations. We represent 3D face geometry in the 2D UV domain, which enables us to train a 2D diffusion model that can incorporate new conditioning modes using IP-Adapters.

\section{Multimodal 3D Face Geometry Generation}
\label{sec:method}

\begin{figure*}
	\centering
	\includegraphics[width=0.9\textwidth]{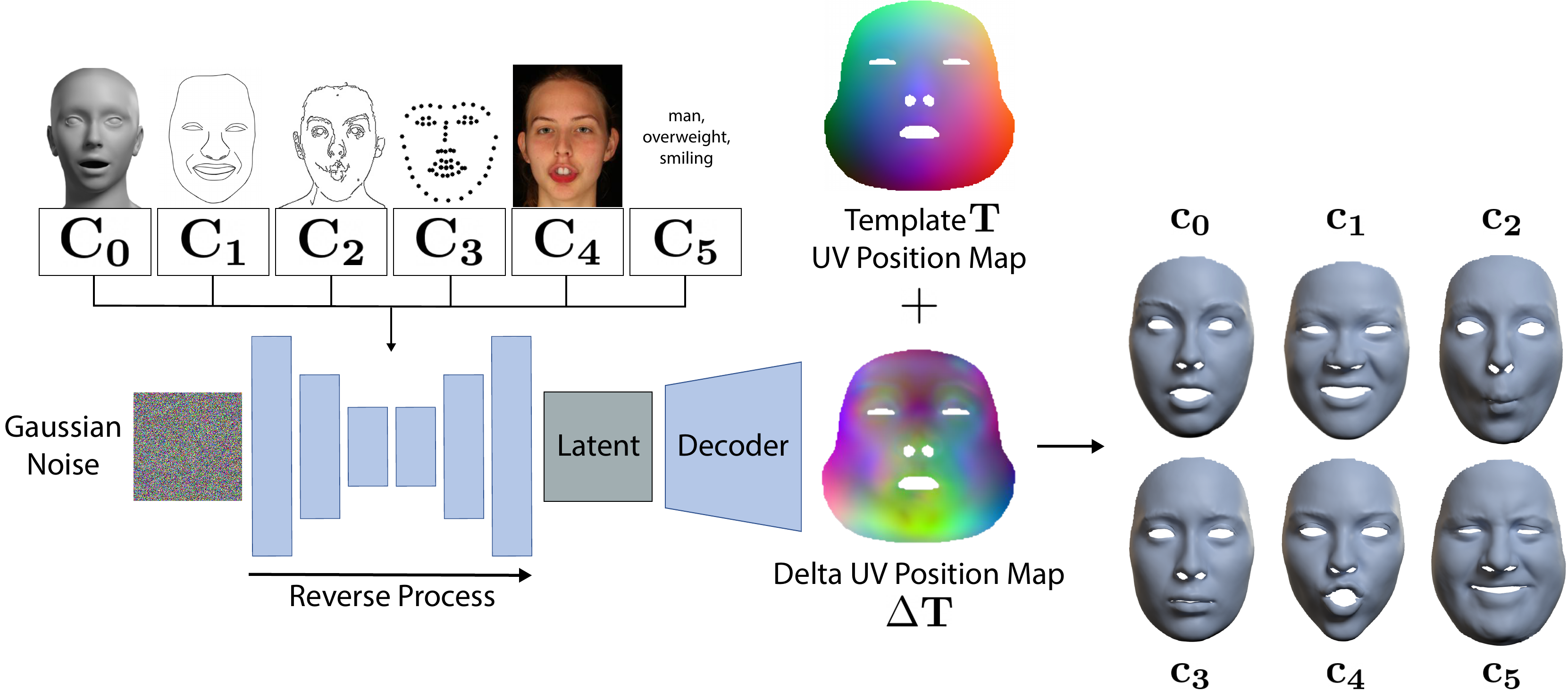}
	\caption{Our pipeline for diffusion-based controllable 3D face geometry generation which uses a delta UV position map representation $\Delta$$\bf{T}$ to generate results. We can control the results with several conditioning modes (i.e. FLAME parameters $\mathbf{c}_0$, sketches $\mathbf{c}_1$, Canny edges $\mathbf{c}_2$, 2D landmarks $\mathbf{c}_3$, portrait photos $\mathbf{c}_4$ and text $\mathbf{c}_5$).
	}
	\label{fig:pipeline}
\end{figure*}

We propose a novel method for diffusion-based controllable 3D face geometry generation that allows for controlling the results with several conditioning modes (i.e. sketches, 2D landmarks, Canny edges, FLAME parameters, portrait photos and text).

Our method consists of four components: First, we create a dataset of 3D faces, where each face is represented as a UV position map describing the vertex positions (\secref{subsec:dataset}).
This data representation can be easily processed with 2D convolutional neural networks. Second, a variational autoencoder (VAE), which compresses our UV position map face data into a latent space representation (\secref{subsec:vae}). Third, a latent diffusion model (LDM), which learns a non-linear, deep controllable face model in latent space (\secref{subsec:ldm}). Fourth, we learn mode-specific cross-attention layers (IP-Adapters) with the ability to transform and inject conditioning modes into the LDM for controllable 3D face geometry generation (\secref{subsec:cond_gen}). Each of the components is explained in the following and visualized in \figref{fig:pipeline}.

\subsection{Dataset and Geometry Representation}
\label{subsec:dataset}

To represent our face geometries, we generate a novel dataset based on the 3D scan data acquired by Chandran \etal~\cite{sdfm}, where all faces are stabilized and in topological correspondence. In total, we use 323 identities in our training dataset, where each identity shows 24 different facial expressions (7752 examples). The full dataset contains 341 identities and we randomly choose around 5\% (all expressions of 18 identities) as a validation set. We subtract a template face shape $\bf{T}$ from all faces in the dataset and thus represent each individual face as a delta from the template face. The computed delta representation $\Delta$$\bf{T}$ reduces artifacts in the generated 3D face geometry, when compared to the full face representation.
We transform each delta face into a vertex delta position map in UV space, which is suitable for being processed by neural networks~\cite{prnet,3dfs,geometry_images}. This representation records the x, y, z coordinates of the face geometry within a 3-channel image, similar to traditional color texture maps, but instead of an RGB value at each pixel we store the x, y, z delta values.
To improve generalization, we augment our existing geometry training data by synthetic identities which we generate via identity interpolation (50k examples) and by mixing face parts of different identities together~\cite{modelingbyexample} (150k examples). Adding the augmentations during LDM training improves the generalization to novel identities when conditioning on the FLAME parameters. We kindly refer to our supplementary material for ablation studies on the geometry representation and the use of geometry data augmentations.
Combining original and augmented data leads to a total training dataset size of 207752 examples. Thus our training dataset is slightly larger in size compared to related 2D image diffusion models that specialize on human faces (\eg~30K images for the CelebA-HQ~\cite{celebahq} face LDM~\cite{ldm}, or the 70K images for the Diffusion Autoencoder in Preechakul \etal~\cite{diffae}), but still much smaller than the datasets required to train general foundation diffusion models that can represent various objects beyond faces. According to Kadkhodaie \etal~\cite{dm_generalization} diffusion models trained on around 100K samples provide evidence of strong generalization in the face domain. We use the parameter space of a common 3D morphable face model (FLAME~\cite{flame}) as a base conditioning because we can fit FLAME to the scan data and to the augmented data and thus generate a large dataset of paired geometry-FLAME parameter data.
Additionally, we create paired training data for several conditioning modes that only have limited paired data available. For example, portrait photos are only available for the scans from the dataset of Chandran \etal~\cite{sdfm}, but not for the augmentation data. However, it is possible to inject new modalities with limited paired data by training new cross attention layers while keeping the LDM frozen as shown by Ye \etal~\cite{ipadapter} (and described in \secref{subsec:cond_gen}).

\subsection{Variational Autoencoder}
\label{subsec:vae}

To reduce the computational requirements for the diffusion model, we downsample our $256^{2}$ UV position map data by a factor of four into a $64^{2}$ latent space using a variational autoencoder (VAE)~\cite{vae} consisting of an encoder $\mathcal{E}$ and a decoder $\mathcal{D}$. We train the VAE from scratch following the autoencoder loss function and architecture as it is presented in related work~\cite{ldm,vqgan}. Specifically, we use the VQ-GAN~\cite{vqgan} autoencoder loss:

\begin{equation}
	\label{eq:vqgan_loss}
	\begin{split}
		\mathcal{L}_\text{VAE} = \mathcal{L}_\text{rec} + \mathcal{L}_\text{GAN} + \mathcal{L}_\text{reg}.
	\end{split}
\end{equation}
$\mathcal{L}_{rec}$ consists of a pixel-wise L1 loss and a LPIPS~\cite{lpips} perceptual loss. It compares the input UV position maps to the reconstructions through the VAE. $\mathcal{L}_{GAN}$ evaluates inputs $x$ and reconstructions $\mathcal{D}(\mathcal{E}(x))$ with a patch-based discriminator~\cite{patchgan} and $\mathcal{L}_{reg}$ employs a codebook loss which serves as a latent space regularizer.  Please refer to Esser \etal~\cite{vqgan} for more details.

\subsection{Latent Diffusion Model}
\label{subsec:ldm}

Next, we train a latent diffusion model (LDM)~\cite{ldm}, that learns to generate latent UV position maps $\mathbf{z} = \mathcal{E}(\mathbf{x})$. To train the LDM, a forward diffusion process is defined as a Markov chain, which noises the latents $\mathbf{z}$ following a fixed noise schedule of $T$ uniformly sampled timesteps. At the last time step $T$, the distribution is Gaussian. We can directly sample $\mathbf{z}_t$ at an arbitrary timestep $t$ by:
\begin{equation}
	\label{eq:forward_process}
	\mathbf{z}_t(\mathbf{z}_0,\bm{\epsilon}) = \sqrt{\bar{\alpha}_t}\mathbf{z}_0 + \sqrt{1-\bar{\alpha}_t}\bm{\epsilon} \hspace{0.5cm} \bm{\epsilon}\sim\mathcal{N}(\mathbf{0},\mathbf{I}),
\end{equation}
where $1-\bar{\alpha}_t$ describes the variance of the noise and $\bar{\alpha}_t:=\prod_{s=1}^{t}\alpha_s$ according to a fixed noise schedule.

We learn to predict the noise $\bm{\epsilon}$ that was added to a noisy latent image $\mathbf{z}_t$ following Ho \etal~\cite{ddpm}:

\begin{equation}
	\label{eq:vqgan_loss}
	\mathcal{L}_{LDM} = \mathop{{}\mathbb{E}_{\mathbf{z}_0,\mathbf{c}_0,t,\bm{\epsilon}\sim \mathcal{N}(\mathbf{0},\mathbf{I})}} \left[ ||\bm{\epsilon} - \bm{\epsilon}_{\theta}(\mathbf{z}_t, \mathbf{c}_0, t)||_2^2 \right],
\end{equation}
where $t$ is the timestep, $\mathbf{c}_0 =  \bm{\rho}_{\phi}(\mathbf{y})$ is a FLAME parameter conditioning and $\bm{\epsilon}_{\theta}(\mathbf{z}_t, \mathbf{c}_0, t)$ is the UNet~\cite{unet} neural network with parameters $\theta$. The FLAME conditioning $\mathbf{c}_0$ is obtained by fitting FLAME to the face geometry encoded in $\mathbf{z_t}$ and mapping it through a MLP $\bm{\rho}_{\phi}(\mathbf{y})$.

During inference (reverse diffusion process) we generate latent 2D UV position maps from the model distribution. We start from $\mathbf{z}_{T}\sim \mathcal{N}(\mathbf{0},\mathbf{I})$ and iteratively compute less noisy latents until we reach a clean latent sample $\mathbf{z}_0$. Sampling following DDPM~\cite{ddpm} or DDIM~\cite{ddim} computes $\mathbf{z}_{t-1}$ from $\mathbf{z}_t$ based on the UNet output.

\subsection{Multimodal Conditional Generation}
\label{subsec:cond_gen}

To control the generations with additional conditioning modes (beyond the FLAME parameters $\mathbf{c}_0$), we train different sets of cross-attention layers, following IP-Adapter~\cite{ipadapter}. The LDM itself is kept frozen. In this way, we can integrate novel conditioning modes post-LDM training even with limited paired mode-geometry data. We train one set for each of the following conditioning modes: sketches $\mathbf{c}_1$, Canny edges $\mathbf{c}_2$, 2D landmarks $\mathbf{c}_3$, portrait photos $\mathbf{c}_4$ and text $\mathbf{c}_5$. The output of the new cross-attention layers is added to the outputs of the existing LDM cross-attention layers, thereby injecting the new conditioning signal into the generation process:
\vspace{-1mm}
\begin{equation}
	\small
	\label{eq:cross_attention}
	\mathbf{Z} = Attention(\mathbf{Q},\mathbf{K},\mathbf{V}) + Attention'(\mathbf{Q},\mathbf{K'}_{m},\mathbf{V'}_{m}),
\end{equation}
where $\mathbf{Q}$ are the intermediate UNet query features, $\mathbf{K}$ and $\mathbf{V}$ are keys and values for our FLAME conditioning $\mathbf{c}_0$ and $\mathbf{K'}_{m}$,$\mathbf{V'}_{m}$ are keys and values for the newly injected modality $\mathbf{c}_m$.
\vspace{-1mm}
\begin{equation}
	\small
	\label{eq:cross_attention_two}
	\begin{aligned}
	\mathbf{K} & = \mathbf{c}_0 \cdot \mathbf{W}_k, \mathbf{V} = \mathbf{c}_0 \cdot \mathbf{W}_v,\\ \mathbf{K'}_{m} & = \mathbf{c}_m \cdot \mathbf{W'}_{k,m}, \mathbf{V'}_{m} = \mathbf{c}_m \cdot \mathbf{W'}_{v,m} 
	\end{aligned}
\end{equation}
Here, $\mathbf{W'}_{k,m}$ and $\mathbf{W'}_{v,m}$ represent the newly added weights that are updated during training.

Prior to passing each of the above-mentioned conditioning modes to the cross-attention layers, we pass each of them through CLIP~\cite{clip} and extract a 768 dimensional global CLIP feature vector, which serves as our conditioning representation. Following IP-Adapter, we train a small projection network consisting of one linear layer and layer normalization, designed to project the CLIP feature vector into several extra context tokens before injecting it into the cross-attention layers. We use 16 tokens for each of our conditionings to allow for meaningful attention computation.

At inference time, we can control the 3D face geometry generation with any of the modes using the respective set of cross-attention layers and classifier-free guidance~\cite{cfg}. The strength of the conditioning signal can be increased by increasing the hyperparameter $w$: 
\vspace{-1mm}
\begin{equation}
	\small
	\label{eq:cfg}
	\hat{\bm{\epsilon}}_{\theta}(\mathbf{z}_t, \mathbf{c}_0, \mathbf{c}_m, t) = w\bm{\epsilon}_{\theta}(\mathbf{z}_t, \mathbf{c}_0, \mathbf{c}_m, t) + (1 - w) \bm{\epsilon}_{\theta}(\mathbf{z}_t, t)
\end{equation}

To condition only on a newly added mode or to generate geometry unconditionally, the FLAME conditioning $\mathbf{c}_0$ is set to its null embedding. Additionally, for unconditional generation, the new cross-attention in \eqnref{eq:cross_attention}, which feeds $\mathbf{c}_m$ to the diffusion model, is simply not added to the original attention. Our full method pipeline is visualized in \figref{fig:pipeline}. For more implementation details, please refer to our supplementary material.

\section{Results}
\label{sec:results}

We now show several results and applications of our new multimodal conditional 3D face geometry generation method. We begin by demonstrating control over the generated facial geometry using FLAME's identity and expression parameters (\secref{subsec:idexpconditioning}). We then demonstrate multimodal conditioned geometry generation by guiding the denoising process using sketches, sparse 2D landmarks, Canny edges, portrait photos and text. We evaluate the effectiveness of these different modalities in guiding the generated geometry in \secref{subsec:multimodalconditioning}. We also compare our method to the state-of-the-art related work both on text and image prompts in \secref{subsec:multimodalconditioning}. In addition to using different conditioning modes, we show how one can also spatially restrict the guidance to a particular face region to perform precise geometry edits in \secref{subsec:geoediting}. We can also generate dynamic facial performances by guiding our model from video inputs and demonstrate that our method can produce facial animations that are stable across time (\secref{subsec:dynamicgeneration}). Finally, we discuss the limitations of our method in \secref{subsec:limitations}.

\subsection{Identity and Expression Conditioning}
\label{subsec:idexpconditioning}

The base conditioning used to train our geometry generator are the identity and expression parameters from the FLAME model~\cite{flame}. We use 300 identity parameters ($\bm{\beta}$), 100 expression parameters ($\bm{\psi}$) and 3 jaw pose ($\bm{\theta}$) parameters.  We combine these FLAME parameters into a 403-dimensional conditioning vector which we pass through a 3-layer MLP with Leaky ReLU activation functions prior to injecting it into the diffusion model via cross-attention. 

Recollect that we do not use the geometry from the FLAME model itself to train our diffusion model. Instead we fit the FLAME model to the high quality facial geometry from Chandran et al.~\cite{sdfm} only to obtain identity and expression parameters, and train the diffusion model directly on the geometry captured by Chandran et al.

\begin{figure}
	\centering
	
\def\imgsize{1.8cm}
\setlength\tabcolsep{2pt}
\def\arraystretch{2}
\begin{tabular}{rccc}
Input \;\;\;\;\, & seed $1$ & seed $2$ & seed $3$\\ \midrule 
\raisebox{-.5\height}{\includegraphics[width=\imgsize]{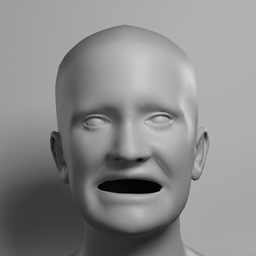}}
& \raisebox{-.5\height}{\includegraphics[width=\imgsize]{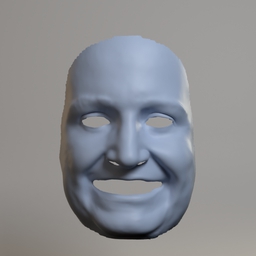}}
& \raisebox{-.5\height}{\includegraphics[width=\imgsize]{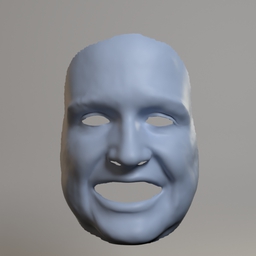}}
& \raisebox{-.5\height}{\includegraphics[width=\imgsize]{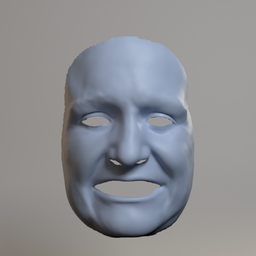}}
\\ 
\raisebox{-.5\height}{\includegraphics[width=\imgsize]{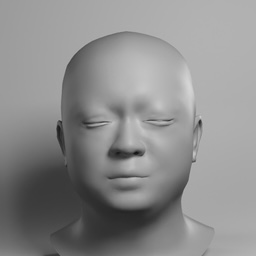}}
& \raisebox{-.5\height}{\includegraphics[width=\imgsize]{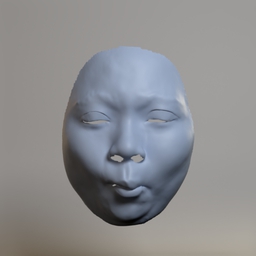}}
& \raisebox{-.5\height}{\includegraphics[width=\imgsize]{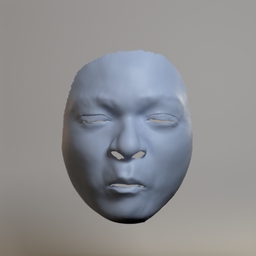}}
& \raisebox{-.5\height}{\includegraphics[width=\imgsize]{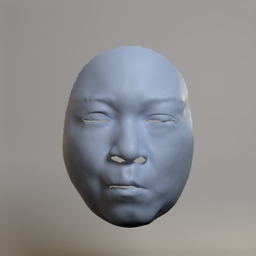}}
\\
\raisebox{-.5\height}{\includegraphics[width=\imgsize]{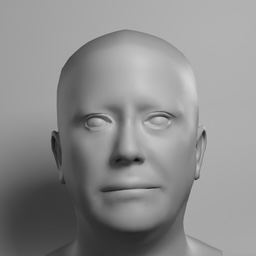}}
& \raisebox{-.5\height}{\includegraphics[width=\imgsize]{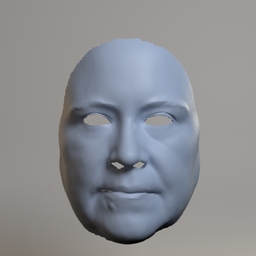}}
& \raisebox{-.5\height}{\includegraphics[width=\imgsize]{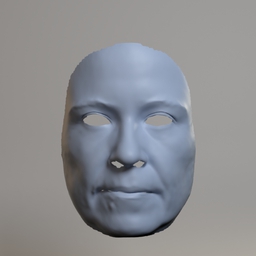}}
& \raisebox{-.5\height}{\includegraphics[width=\imgsize]{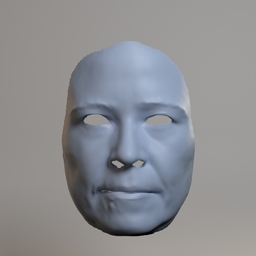}}

\end{tabular}

	\caption{Changing the noise seed while conditioning on the FLAME parameters does not affect the identity and expression of the generated geometry, but only the stochastic details that are added on top. Our model can capture richer geometric detail that is not present in the original FLAME mesh, while still respecting FLAME's identity and expression parameters. Each row shows a different set of FLAME parameters, with the corresponding FLAME mesh visualized in the first column.}
	\label{fig:stochasticdetails}
\end{figure}

We visualize geometries generated by our model for unseen FLAME parameters in ~\figref{fig:stochasticdetails}. As our underlying mesh topology is different from FLAME and represents $\sim$10-times more vertices, it can express a greater level of detail that is not present in the lower resolution FLAME mesh. This high resolution detail is captured and reproduced by the denoising process. Therefore, by simply varying the noise seed, one can obtain variations of the FLAME-conditioned geometry, each of which contain different mid/high frequency details.

\begin{figure}
	\centering
	\includegraphics[width=1.0\columnwidth]{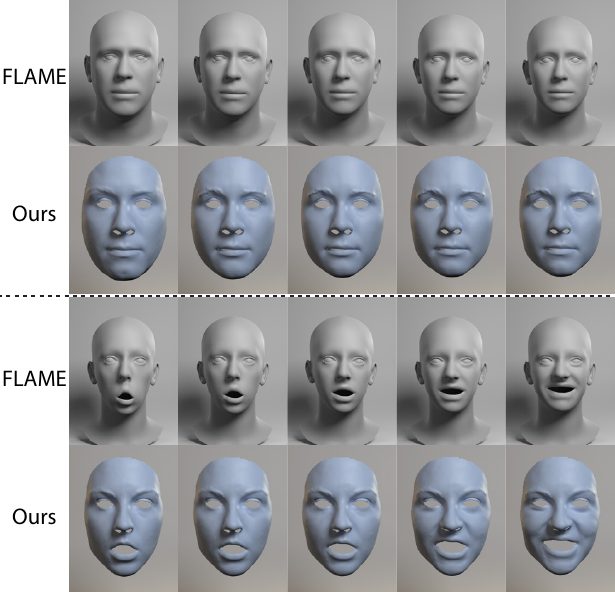}
	\caption{Identity and expression disentanglement. Traversing the first dimension in FLAME's identity space leads to smooth changes in our generated face geometry (rows 1 and 2). Similarly a linear interpolation of FLAME's expression parameters results in a smooth, yet nonlinear, interpolation of facial expression as seen in rows 3 and 4. Please refer to our supplementary video to see the complete interpolation.}
	\label{fig:interpolation}
\end{figure}

\noindent{\textbf{Disentangling Identity and Expression.}}
Our diffusion model also preserves the disentanglement between facial identity and expression that is present in FLAME. In \figref{fig:interpolation}, we show how a smooth interpolation of FLAME's identity and expression coefficients results in a smooth, yet nonlinear, interpolation of our generated geometry.  We can observe that the facial expression remains fixed when interpolating between identities, and vice versa (please also refer to the supplemental video). To eliminate the randomness in the generation, we used the same initial noise to generate the interpolated geometries along with DDIM sampling~\cite{ddim}.

\subsection{Multimodal Conditioning}
\label{subsec:multimodalconditioning}

\begin{figure}
	\centering
	
\def\imgsize{1.6cm}
\setlength\tabcolsep{1.5pt}
\def\arraystretch{1.5}
\begin{tabular}{rcccc}
Portrait
& \raisebox{-.5\height}{\includegraphics[width=\imgsize]{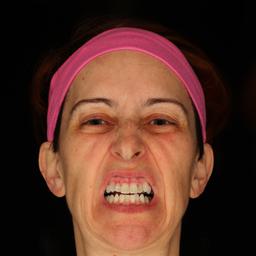}}
& \raisebox{-.5\height}{\includegraphics[width=\imgsize]{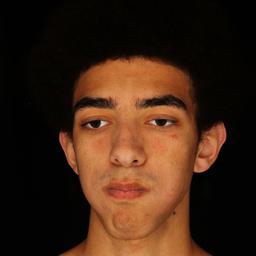}}
& \raisebox{-.5\height}{\includegraphics[width=\imgsize]{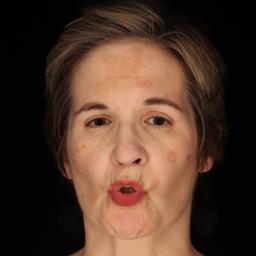}}
& \raisebox{-.5\height}{\includegraphics[width=\imgsize]{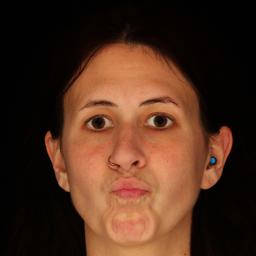}}
 \\
Output
& \raisebox{-.5\height}{\includegraphics[width=\imgsize]{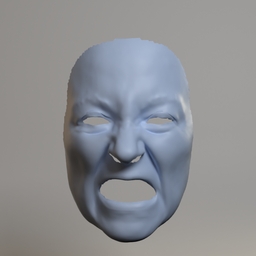}}
& \raisebox{-.5\height}{\includegraphics[width=\imgsize]{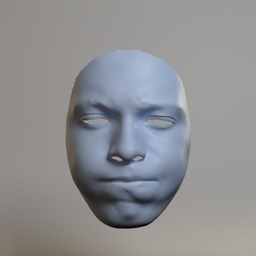}}
& \raisebox{-.5\height}{\includegraphics[width=\imgsize]{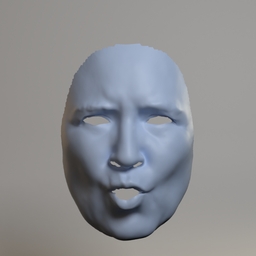}}
& \raisebox{-.5\height}{\includegraphics[width=\imgsize]{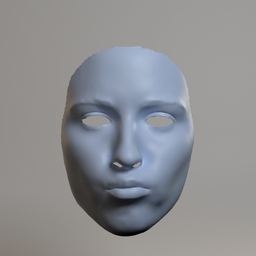}}
\\ \midrule 
Sketch
& \raisebox{-.5\height}{\includegraphics[width=\imgsize]{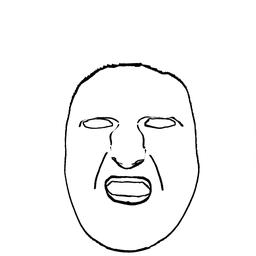}}
& \raisebox{-.5\height}{\includegraphics[width=\imgsize]{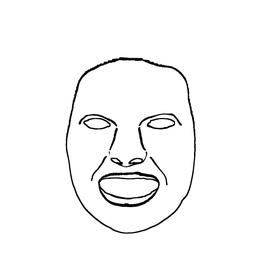}}
& \raisebox{-.5\height}{\includegraphics[width=\imgsize]{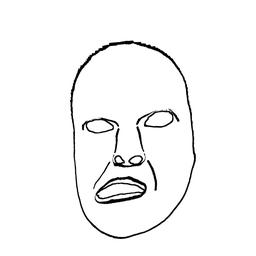}}
& \raisebox{-.5\height}{\includegraphics[width=\imgsize]{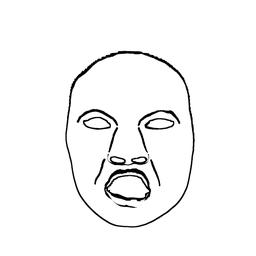}}
 \\
Output
& \raisebox{-.5\height}{\includegraphics[width=\imgsize]{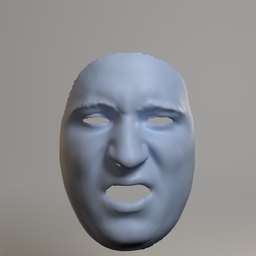}}
& \raisebox{-.5\height}{\includegraphics[width=\imgsize]{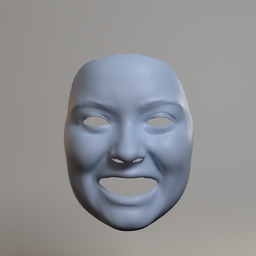}}
& \raisebox{-.5\height}{\includegraphics[width=\imgsize]{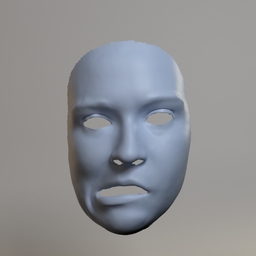}}
& \raisebox{-.5\height}{\includegraphics[width=\imgsize]{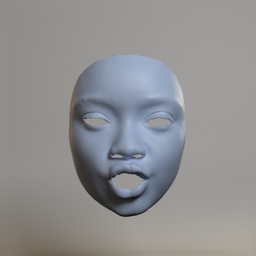}}
\\ \midrule 
FLAME
& \raisebox{-.5\height}{\includegraphics[width=\imgsize]{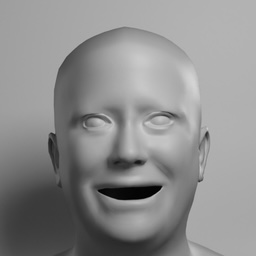}}
& \raisebox{-.5\height}{\includegraphics[width=\imgsize]{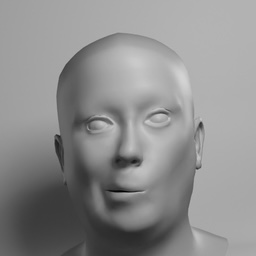}}
& \raisebox{-.5\height}{\includegraphics[width=\imgsize]{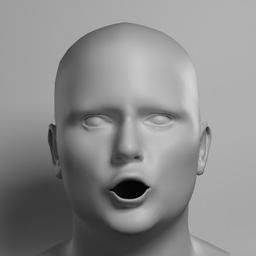}}
& \raisebox{-.5\height}{\includegraphics[width=\imgsize]{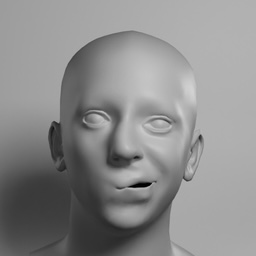}}
 \\
Output
& \raisebox{-.5\height}{\includegraphics[width=\imgsize]{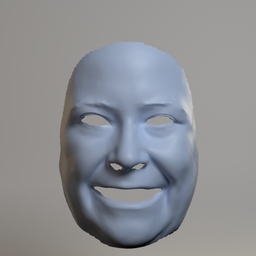}}
& \raisebox{-.5\height}{\includegraphics[width=\imgsize]{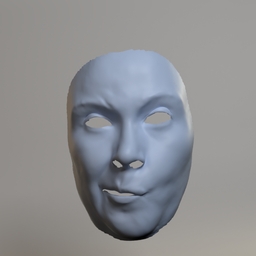}}
& \raisebox{-.5\height}{\includegraphics[width=\imgsize]{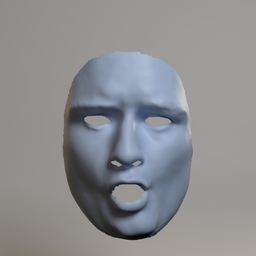}}
& \raisebox{-.5\height}{\includegraphics[width=\imgsize]{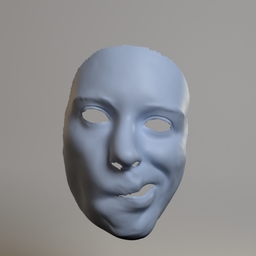}}
\\ \midrule 
Edges
& \raisebox{-.5\height}{\includegraphics[width=\imgsize]{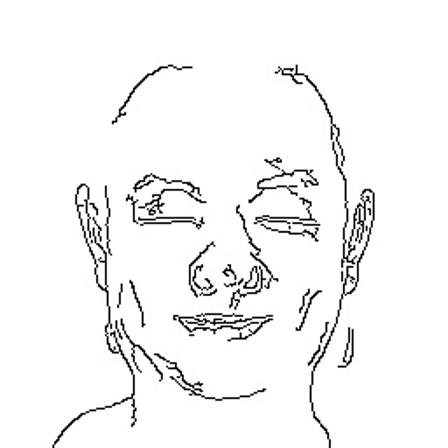}}

& \raisebox{-.5\height}{\includegraphics[width=\imgsize]{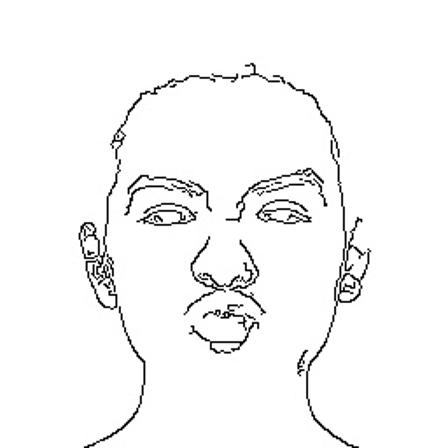}}
& \raisebox{-.5\height}{\includegraphics[width=\imgsize]{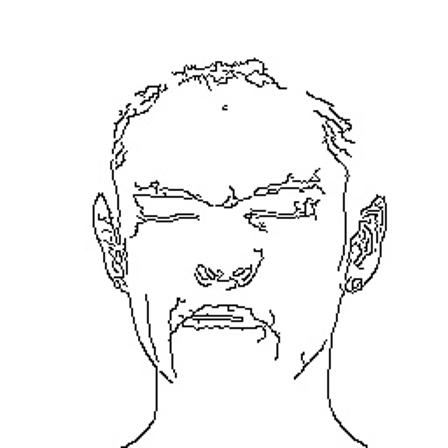}}
& \raisebox{-.5\height}{\includegraphics[width=\imgsize]{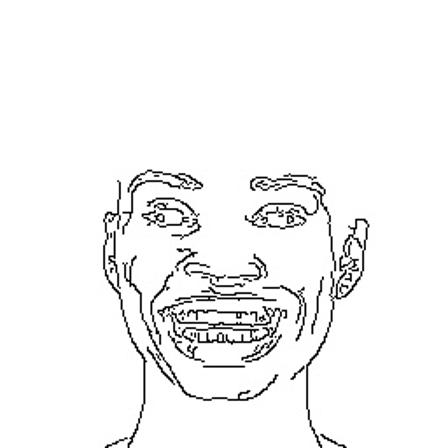}}
 \\
Output
& \raisebox{-.5\height}{\includegraphics[width=\imgsize]{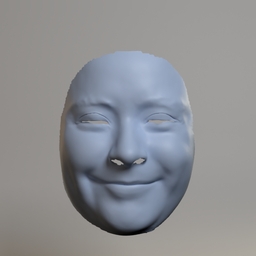}}
& \raisebox{-.5\height}{\includegraphics[width=\imgsize]{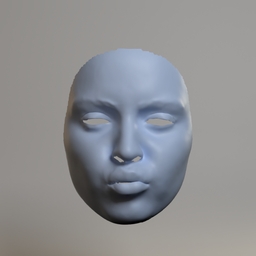}}
& \raisebox{-.5\height}{\includegraphics[width=\imgsize]{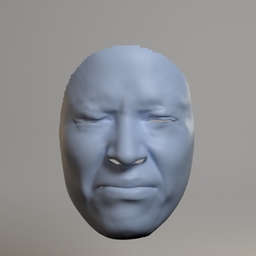}}
& \raisebox{-.5\height}{\includegraphics[width=\imgsize]{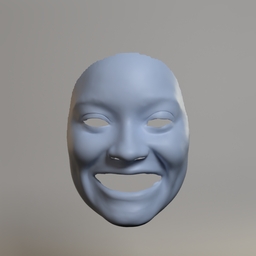}}

\\ \midrule 
Landmarks
& \raisebox{-.5\height}{\includegraphics[width=\imgsize]{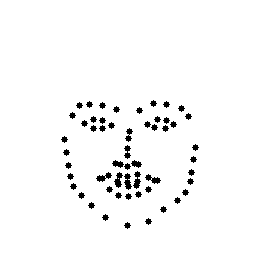}}
& \raisebox{-.5\height}{\includegraphics[width=\imgsize]{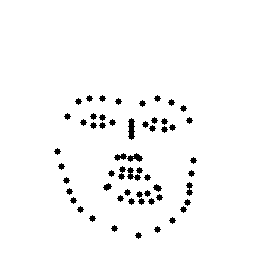}}
& \raisebox{-.5\height}{\includegraphics[width=\imgsize]{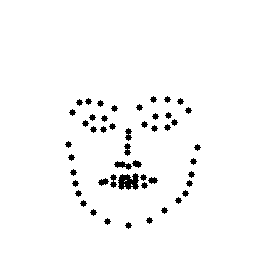}}
& \raisebox{-.5\height}{\includegraphics[width=\imgsize]{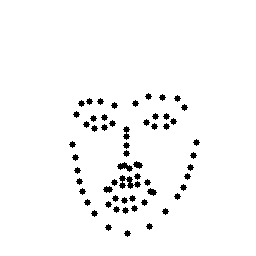}}

 \\
Output
& \raisebox{-.5\height}{\includegraphics[width=\imgsize]{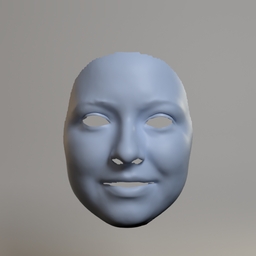}}
& \raisebox{-.5\height}{\includegraphics[width=\imgsize]{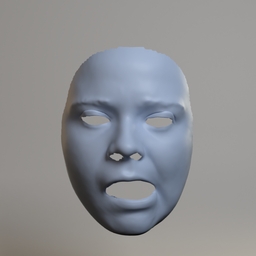}}
& \raisebox{-.5\height}{\includegraphics[width=\imgsize]{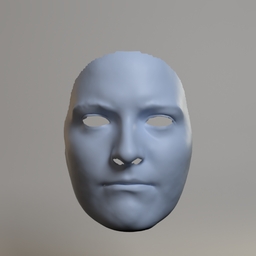}}
& \raisebox{-.5\height}{\includegraphics[width=\imgsize]{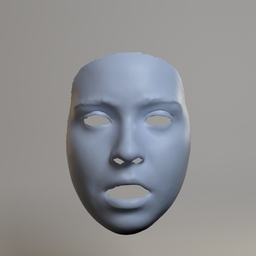}}
\\ \midrule 
Text
& \raisebox{-.5\height}{\includegraphics[width=\imgsize]{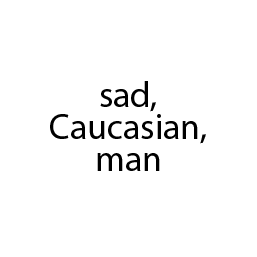}}
& \raisebox{-.5\height}{\includegraphics[width=\imgsize]{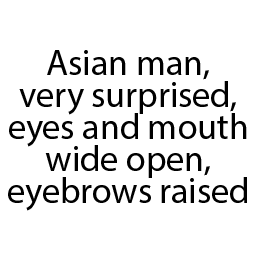}}
& \raisebox{-.5\height}{\includegraphics[width=\imgsize]{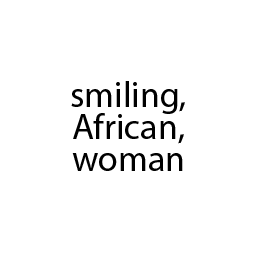}}
& \raisebox{-.5\height}{\includegraphics[width=\imgsize]{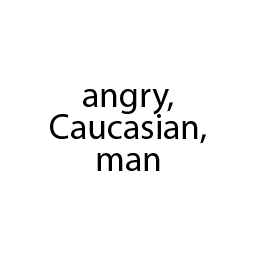}}

 \\
Output
& \raisebox{-.5\height}{\includegraphics[width=\imgsize]{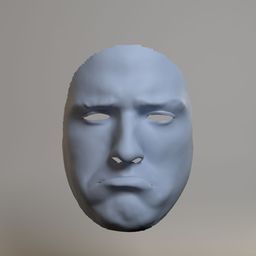}}
& \raisebox{-.5\height}{\includegraphics[width=\imgsize]{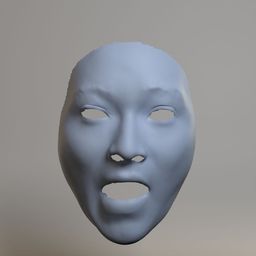}}
& \raisebox{-.5\height}{\includegraphics[width=\imgsize]{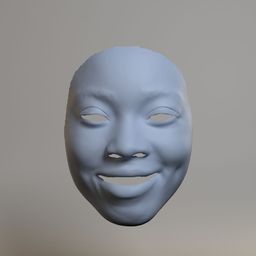}}
& \raisebox{-.5\height}{\includegraphics[width=\imgsize]{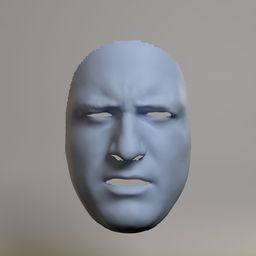}}
\end{tabular}

	\caption{Multimodal conditional generation results on our validation dataset, including conditioning on portrait images, sketches, FLAME parameters, Canny edges, 2D landmarks and text. The FLAME parameter inputs are visualized as meshes. Our model captures the facial identity and expression across all conditioning modes. We use classifier-free guidance of $w=1$, except for some text-prompts where we use $w=3$ for even stronger facial expressions.}
	\label{fig:validation-modes}
\end{figure}

Beyond the underlying FLAME-based control, we introduce additional conditioning modes to control our diffusion model following \secref{subsec:cond_gen}. We now discuss the results of facial geometry generation by conditioning our diffusion model on sketches, sparse 2D landmarks, Canny edges, portrait RGB photos and text. Geometries generated for different conditionings from each of these additional modes are shown in \figref{fig:validation-modes}. 

\begin{table}
	\caption{\label{tab:v2v} We report the vertex-to-vertex error (V2V) to the original scanned geometry in mm on our validation set of 432 shapes for generations from different types of conditioning signals. Conditions that are more descriptive of the end facial geometry like FLAME parameters, and portrait images achieve a lower error than others. Results are averaged over three different seeds.} 
	\centering
	\begin{tabular}{l c c c}
		\toprule
		\textbf{V2V error} & \textbf{Mean $\downarrow$} & \textbf{Median $\downarrow$} & \textbf{Std $\downarrow$}  \\
		\midrule
		2D landmarks & 6.390 & 5.950 & 1.945 \\
		Canny edges
		& 6.007 &  5.701 &  1.672 \\
		Sketch
		& 5.521 & 5.106 & 1.792 \\
		Portrait photo
		& 5.207 & 4.942 & \textbf{1.471} \\
		FLAME parameters
		& \textbf{5.008} & \textbf{4.737} & 1.498 \\
		\midrule
	\end{tabular}
\end{table}

\begin{table}
	\caption{\label{tab:sota_comp_text} We report the CLIP score (ViT-B/32) for \textbf{text}-to-geometry generation to related work. We prompt each method with 10 text prompts specifying a neutral expression. Additionally, we compare the average inference-time per sample between different methods on a single 3090 GPU. *Describe3D inference-time is measured on a single 1080 GPU and DreamFace-V2 results are exported from their web interface (N/A). We run \textit{FaceG2E fast} for only 15/200 optimization steps to match the inference-time of our method for comparison. The best score per column is marked in bold and the second-best score is underlined.}
	\centering
	\resizebox{\columnwidth}{!}{
	\begin{tabular}{l c c c}
		\toprule
		\textbf{Method} & \textbf{CLIP (ViT-B/32) $\uparrow$} & \textbf{Time $\downarrow$}  \\
		\midrule
		Describe3D~\cite{describe3d} & 32.69 & 80.62* \\
		DreamFace-V2~\cite{dreamface}
		& 33.85 &  N/A \\
		FaceG2E fast~\cite{faceg2e}
		& 32.41 & $\underline{6.07}$ \\
		FaceG2E~\cite{faceg2e}
		& \textbf{35.03} & 42.87 \\
		Ours
		& $\underline{34.15}$ & \textbf{5.48} \\
		\midrule
	\end{tabular}
}
\end{table}

\noindent{\textbf{Quantitative Comparison.}}
While our generated geometry follows the identity and expression seen in the input conditioning signal, the degree to which the conditioning signal constrains the generated geometry varies from mode to mode. We evaluate the effectiveness of each of our conditioning modes in guiding the generated geometry towards ground truth scans, by computing the Euclidean error between the generated geometry and the ground truth geometry for each conditioning mode. In \tabref{tab:v2v}, we report the vertex-to-vertex (V2V) error of 432 shapes from our validation set for each type of conditioning. We observe that conditioning signals that are more descriptive, such as FLAME parameters or portrait photos, obtain a lower error when compared to signals that are less descriptive of the final geometry (2D landmarks, Canny edges and sketches). Please refer to our supplementary material for error maps on our validation set.

\begin{figure}
	\centering
	\includegraphics[width=1.0\columnwidth]{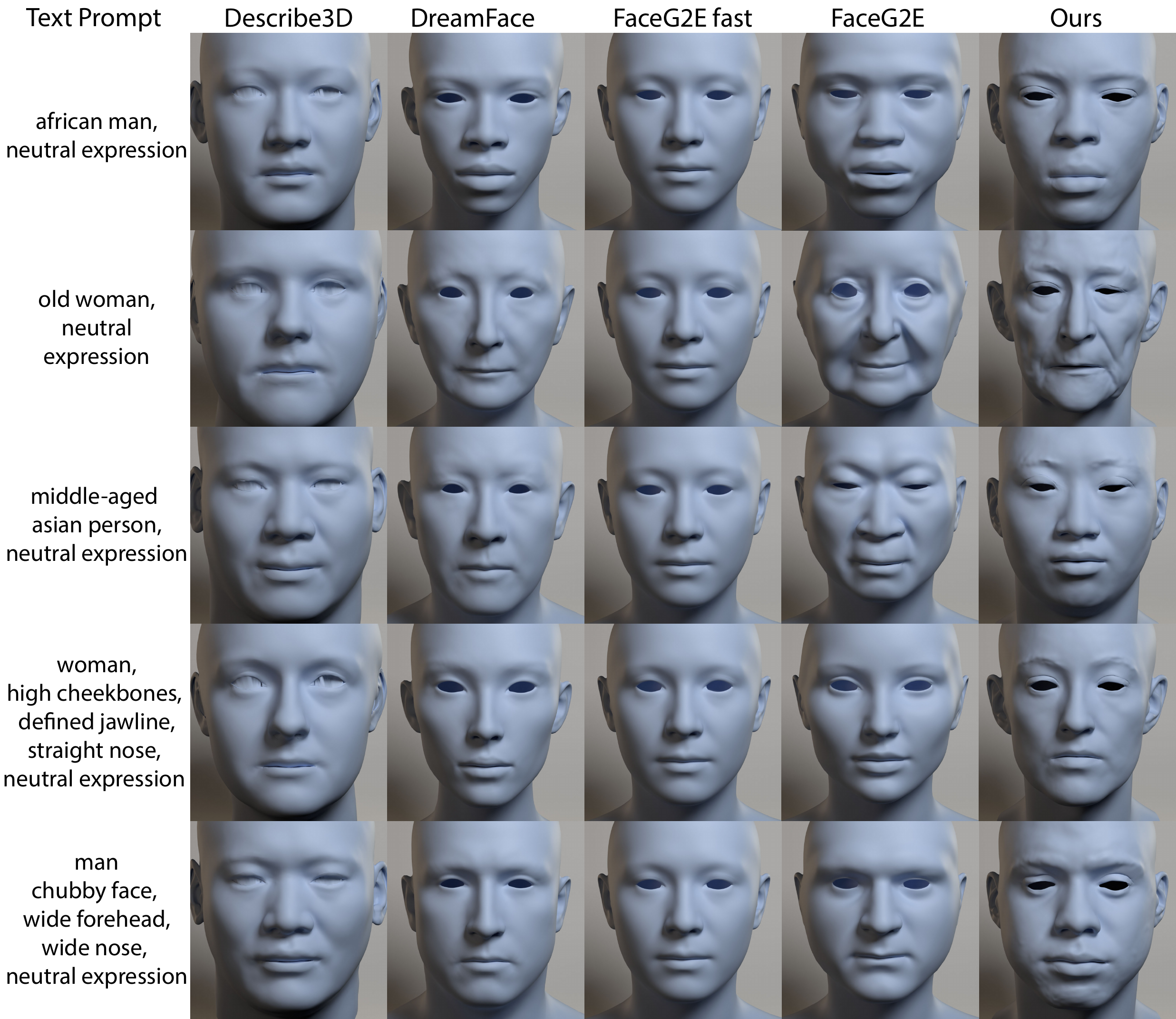}
	\caption{Comparison to related work on text-to-geometry generation with neutral expression prompts. We sample from our model using classifier-free guidance $w=3$.}
	\label{fig:sota_comparison_neutral_figure}
\end{figure}

\begin{figure}
	\centering
	\includegraphics[width=1.0\columnwidth]{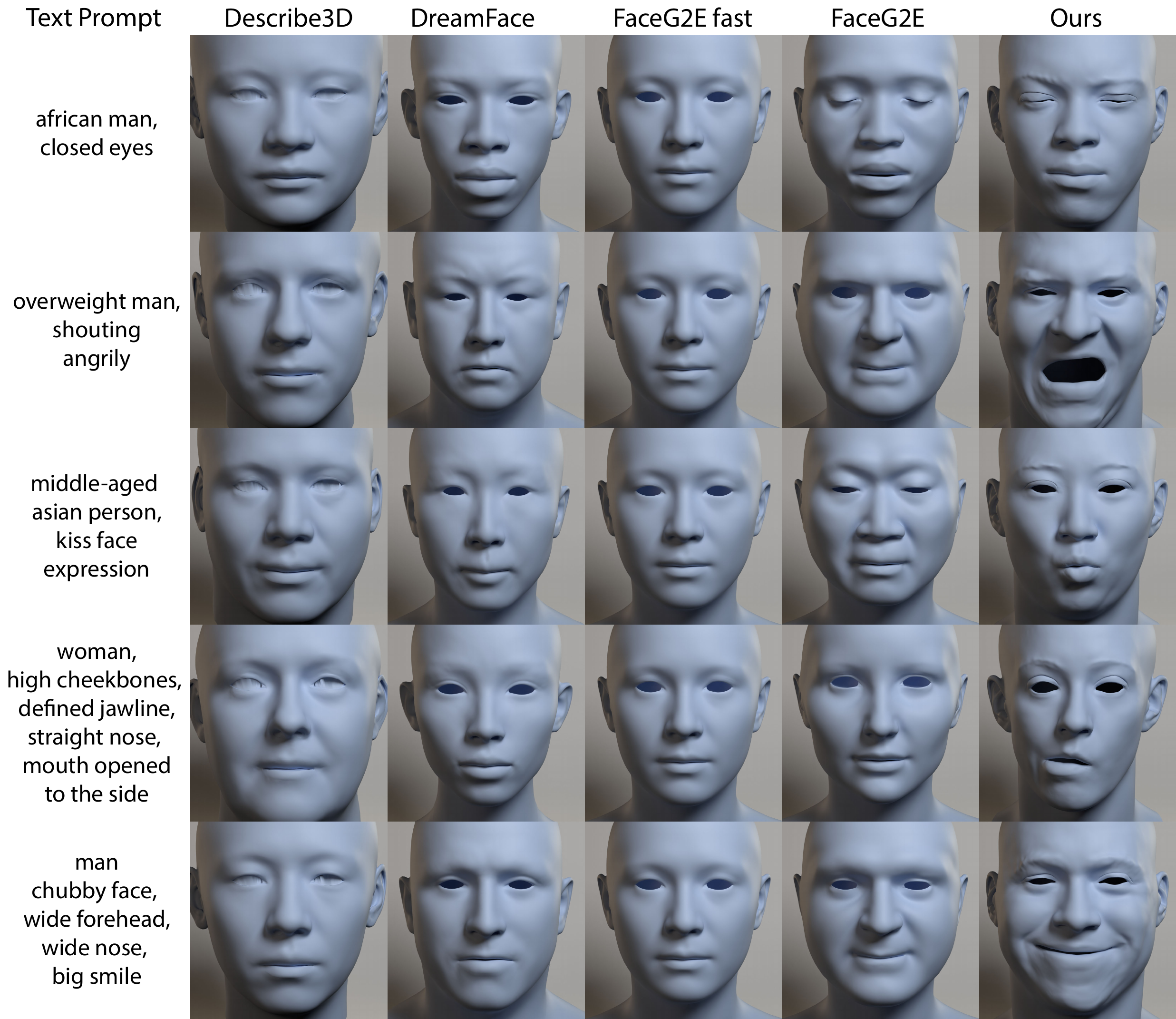}
	\caption{Comparison to related work on text-to-geometry generation with expression prompts.}
	\label{fig:sota_comparison_expression_figure}
\end{figure}

Next, we compare our method to the state-of-the-art related work methods. First, we look at the text conditioning mode, because prompts from this mode are most commonly supported by related work. We compare our method to Describe3D~\cite{describe3d}, DreamFace~\cite{dreamface} and two variants of FaceG2E~\cite{faceg2e}, over 10 different text prompts. We report the average CLIP score (ViT-B/32) between the CLIP embeddings from the text prompt and the CLIP embeddings extracted from the rendered generated face geometry. Specifically, we evaluate text prompts that specify varied identities all with a neutral expression, because neutrals are supported by all methods. Each prompt is prepended with "\emph{A shaded, textureless 3D face model of}", although for legibility, we shorten the text prompts when we add them to figures (\eg~\figref{fig:sota_comparison_neutral_figure} and \figref{fig:sota_comparison_expression_figure}). Please refer to our supplementary material for the exact text prompts. Among the compared methods, our method has the second highest CLIP score on text prompts that specify a neutral expression (Table~\ref{tab:sota_comp_text}). 

\noindent{\textbf{Qualitative Comparison.}}  We first show qualitative comparisons to state-of-the-art text-to-geometry methods on {\em neutral} text prompts in ~\figref{fig:sota_comparison_neutral_figure}. Our method produces realistic faces that are subjectively on par with other techniques.  Furthermore, it is important note that our method also natively supports generating faces with {\em expressions} by providing an expression description in the text prompt. This is a situation that other methods struggle with, as shown in \figref{fig:sota_comparison_expression_figure}.  

\begin{figure}
	\centering
	\includegraphics[width=0.68\columnwidth]{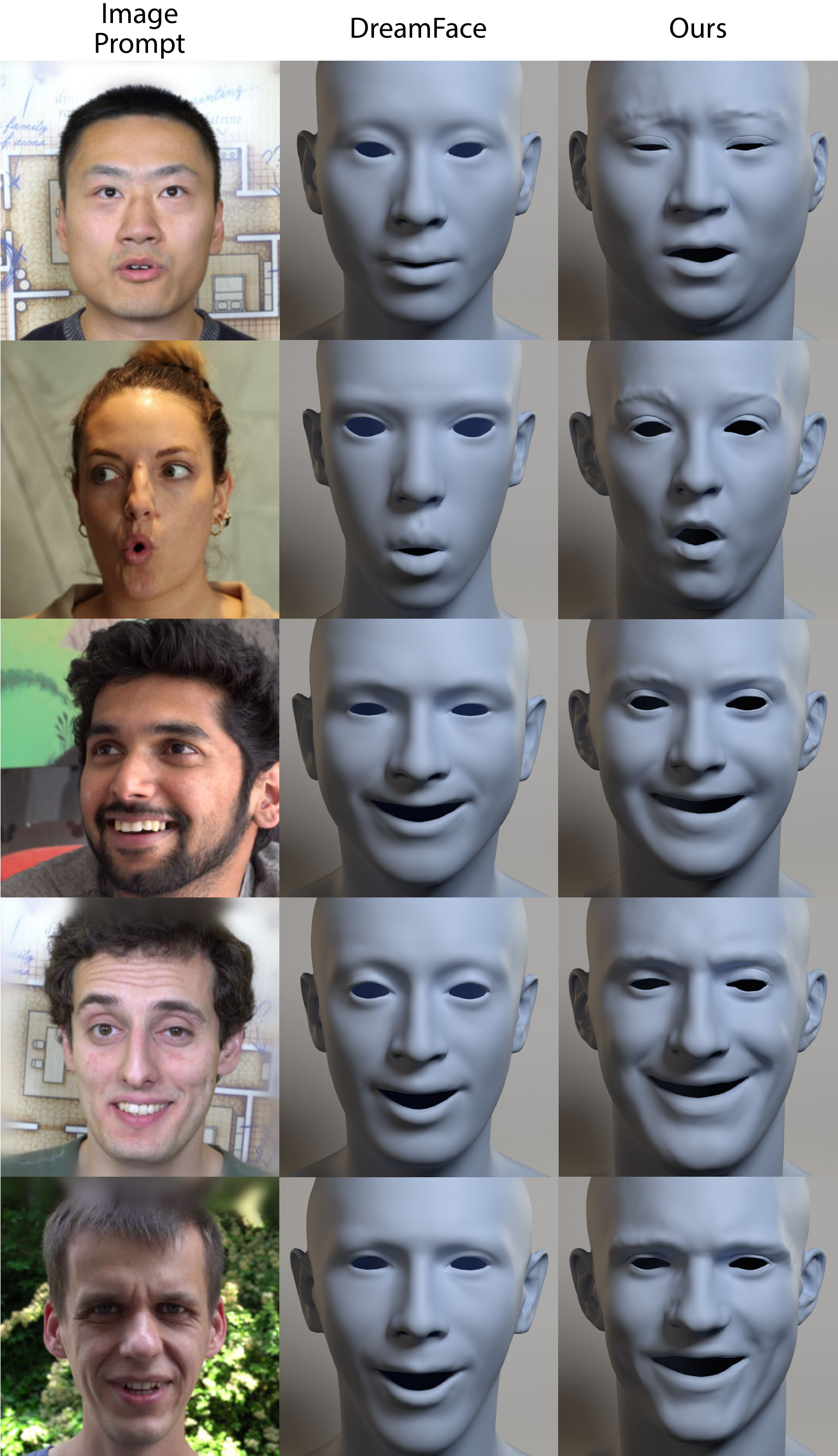}
	\caption{Image prompt to face geometry generation. Comparison between DreamFace~\cite{dreamface} and our method on in-the-wild test data.}
	\label{fig:sota_comparison_image_prompt}
\end{figure}

\begin{figure}
	\centering
	\includegraphics[width=1.0\columnwidth]{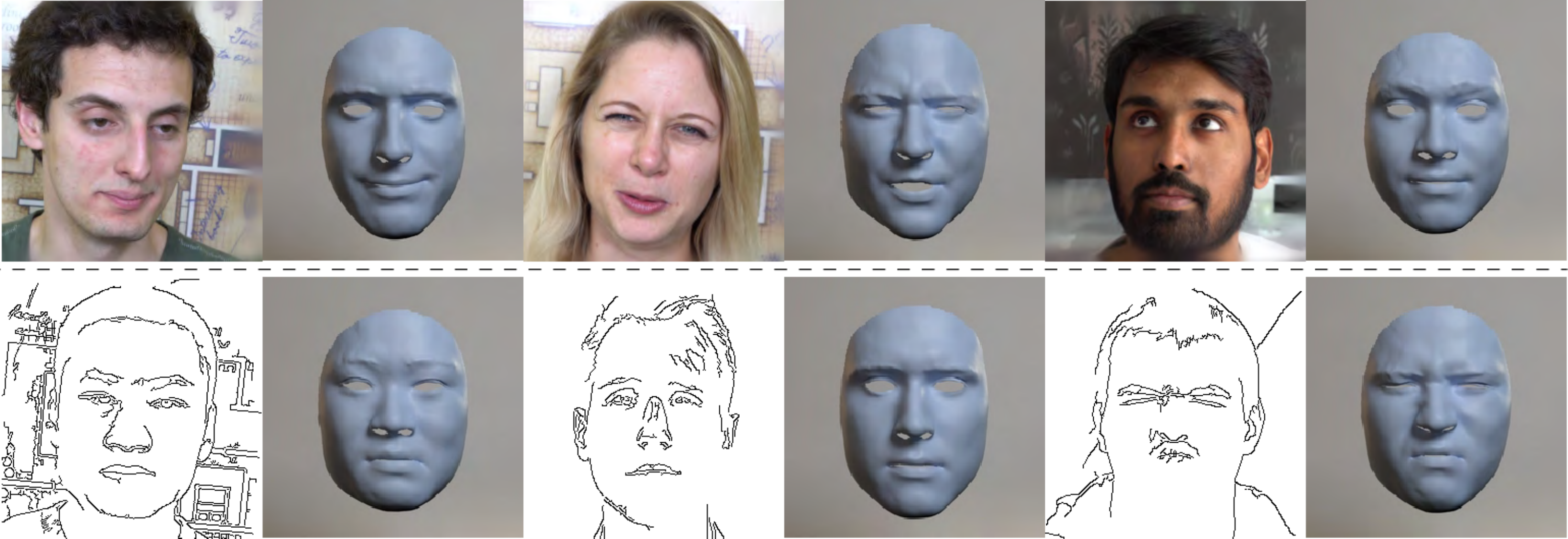}
	\caption{Generation using conditioning signals obtained from in-the-wild test data (Portrait images top row, Canny edge maps bottom row). Our model produces reasonable facial geometry from in-the-wild conditions despite being trained only on studio data.}
	\label{fig:inthewild}
\end{figure}

Second, besides text prompts, DreamFace and our method also support RGB image prompts (\figref{fig:sota_comparison_image_prompt}). Our method achieves results comparable to DreamFace without relying on SDS optimization as part of the geometry generation. Note, that to aid the visual comparisons with previous methods, we complete the head in our results by deforming a template head to match our generated face. 
Despite training on purely studio data, we show how our model responds to conditionings derived from in-the-wild data in \figref{fig:inthewild}. For direct visual comparison with the identity and facial expression in the conditioning signal, we overlay the generated meshes onto validation images from whom the respective conditionings were extracted in \figref{fig:mesh_overlay}.

\begin{figure}
	\centering
	\includegraphics[width=0.8\columnwidth]{figures/mesh_overlay/mesh_overlay.jpg}
	\caption{We overlay our generated meshes on top of the images that display the identities (and expressions) from whom the respective conditioning signals were extracted. We show results for FLAME and portrait photo conditioning.}
	\label{fig:mesh_overlay}
\end{figure}

\begin{figure}[h]
	\includegraphics[width=1.0\columnwidth]{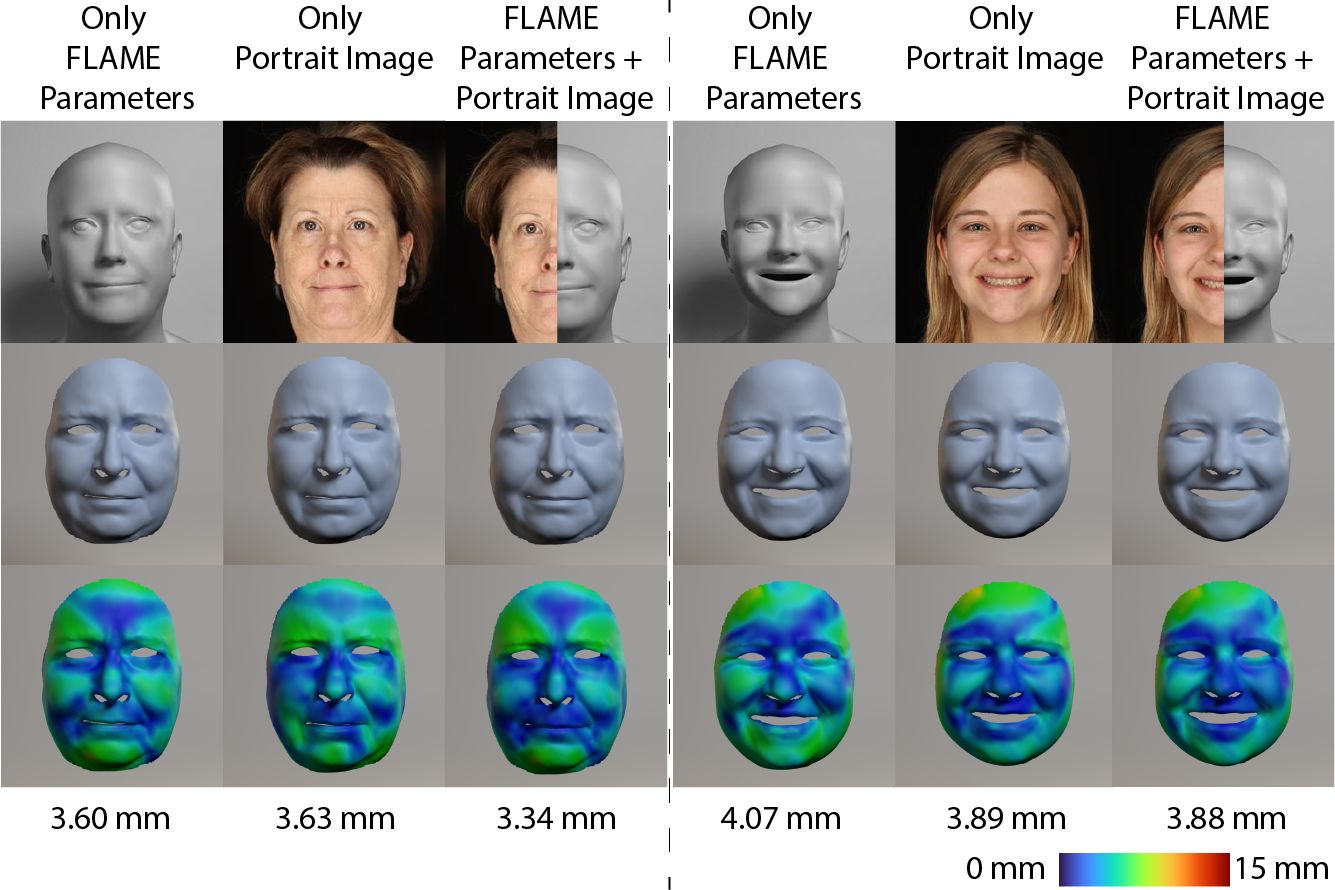}
	\caption{Multimodal conditioning using portrait photo and FLAME parameter conditioning separately and simultaneously. The first row shows the conditioning inputs. The second row shows the generated face geometry. The third row shows the error map when compared with the original geometry from our validation set.}
	\label{fig:multimodalprompt}
\end{figure}

\noindent{\textbf{Combining Two Conditionings.}}
As our method involves learning additional modes of conditioning on top of an underlying FLAME conditioned diffusion model, we can also use more than one conditioning signal at inference time to guide the generation. In \figref{fig:multimodalprompt}, we show how combining both FLAME parameter and portrait image conditioning lowers the vertex error on a validation sample, as the denoising UNet now has access to more information about the desired identity and expression. Note that only our base conditioning (FLAME parameters) is present when training any one of the other modalities cross-attention layers. Therefore, we can expect complementary results only when combining the FLAME parameter conditioning with another modality.

\noindent{\textbf{Inference time.}} We show that our method has significant inference-time benefits over its competitors that are mainly based on SDS optimization. It has the fastest average inference speed per sample on text-to-geometry generation, because once it is trained, it can directly sample 3D faces from the diffusion model. This setup alleviates the need for compute intensive iterative SDS optimization of a 3D face representation. Specifically, our method is more than seven times faster than its closest competitor FaceG2E (5.48 seconds vs. 42.87 seconds). To compare results with similar inference speeds, we run \textit{FaceG2E fast} for approximately the same time as our own method and evaluate the CLIP score results (see Table~\ref{tab:sota_comp_text}).

\begin{figure}
	
\def\imgsize{1.6cm}
\setlength\tabcolsep{2pt}
\def\arraystretch{1.5}
\begin{tabular}{rcccc}
Mode & Input & $w=0$ & $w=1$ & $w=3$\\ \midrule 
Portrait
& \raisebox{-.5\height}{\includegraphics[width=\imgsize]{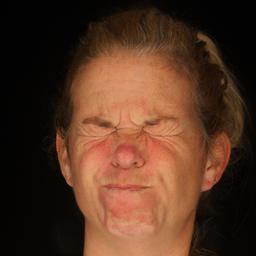}}
& \raisebox{-.5\height}{\includegraphics[width=\imgsize]{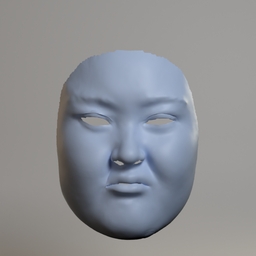}}
& \raisebox{-.5\height}{\includegraphics[width=\imgsize]{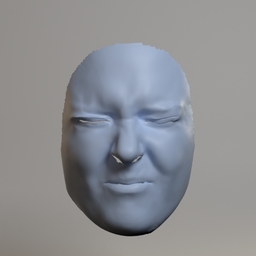}}
& \raisebox{-.5\height}{\includegraphics[width=\imgsize]{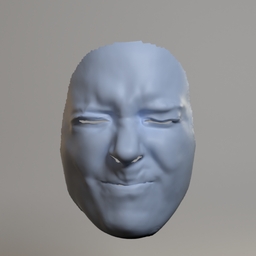}}
\\ 
Sketch
& \raisebox{-.5\height}{\includegraphics[width=\imgsize]{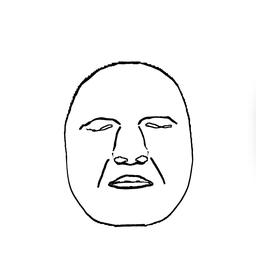}}
& \raisebox{-.5\height}{\includegraphics[width=\imgsize]{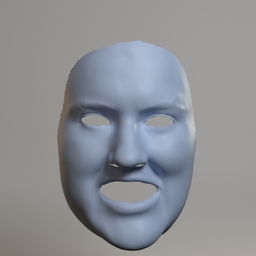}}
& \raisebox{-.5\height}{\includegraphics[width=\imgsize]{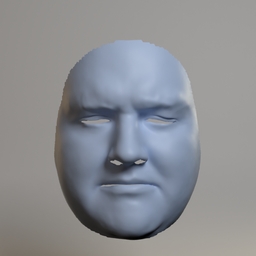}}
& \raisebox{-.5\height}{\includegraphics[width=\imgsize]{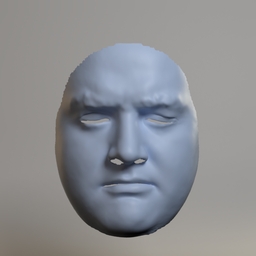}}
\\ 
Flame
& \raisebox{-.5\height}{\includegraphics[width=\imgsize]{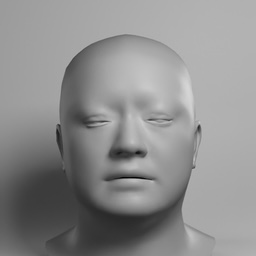}}
& \raisebox{-.5\height}{\includegraphics[width=\imgsize]{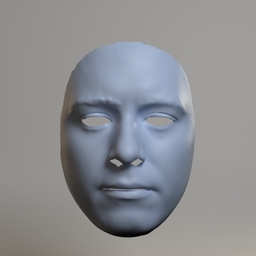}}
& \raisebox{-.5\height}{\includegraphics[width=\imgsize]{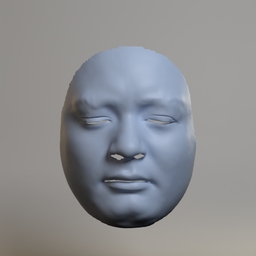}}
& \raisebox{-.5\height}{\includegraphics[width=\imgsize]{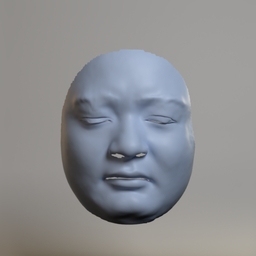}}
\\ 
Edges
& \raisebox{-.5\height}{\includegraphics[width=\imgsize]{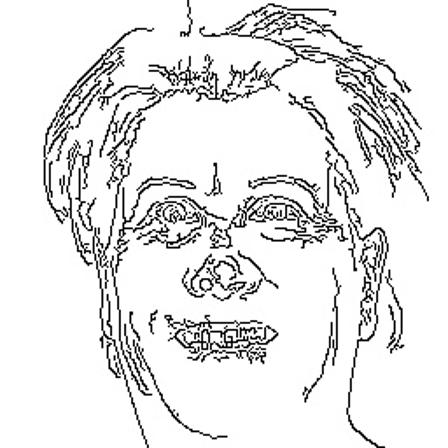}}
& \raisebox{-.5\height}{\includegraphics[width=\imgsize]{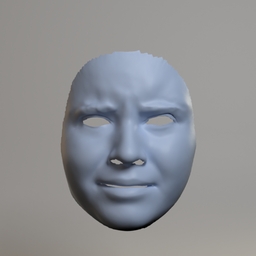}}
& \raisebox{-.5\height}{\includegraphics[width=\imgsize]{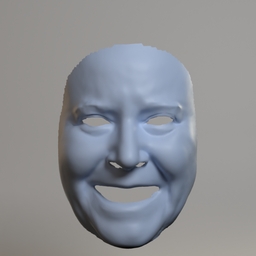}}
& \raisebox{-.5\height}{\includegraphics[width=\imgsize]{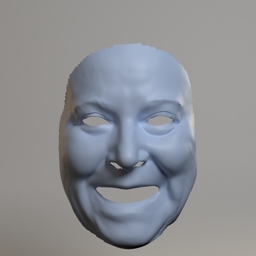}}
\\ 
Landmarks
& \raisebox{-.5\height}{\includegraphics[width=\imgsize]{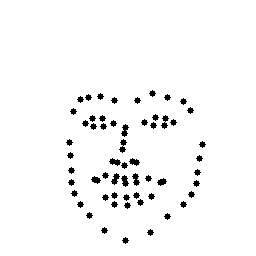}}
& \raisebox{-.5\height}{\includegraphics[width=\imgsize]{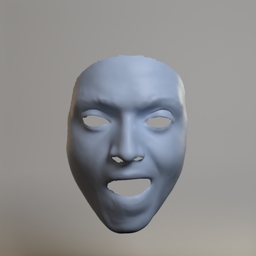}}
& \raisebox{-.5\height}{\includegraphics[width=\imgsize]{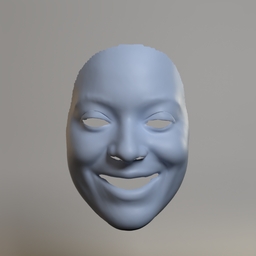}}
& \raisebox{-.5\height}{\includegraphics[width=\imgsize]{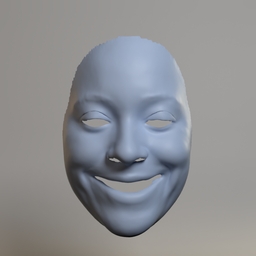}}
\end{tabular}

	\caption{By varying the guidance strength $w$, we can control the extent to which our conditioning signals affect the generated geometry. Setting $w = 0$ results in unconditional generation, while $w >= 1$ results in conditional generation.}
	\label{fig:guidance}
\end{figure}

\noindent{\textbf{Controlling the Conditioning Strength.}}
To control the strength $w$ of the guiding condition, we make use of classifier-free guidance~\cite{cfg} following \eqnref{eq:cfg}. Increasing the guidance strength increases the effect that the input conditioning (prompt) has on the resulting geometry. Stronger guidance can lead to increased level of detail in the generated face geometry and greater resemblance with the input prompt. For example, in \figref{fig:guidance}, the expression of the generated geometry of the subject in the first row displays stronger wrinkles, and a closer match to the portrait image when setting $w=3$ compared to setting it to $w=1$. Unless specified differently, we use $w=1$ for all our conditional generation results.

\subsection{Geometry Editing}
\label{subsec:geoediting}
\begin{figure}
	\centering
	\setlength\tabcolsep{0pt}
\def\arraystretch{1.2}
\begin{tabular}{ccccc}
    \begin{overpic}[width=0.2\linewidth,trim={0 0 0 10pt},clip]{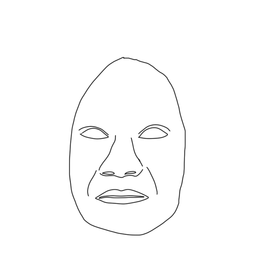}
    \end{overpic}
     &
    \begin{overpic}[width=0.2\linewidth,trim={0 0 0 10pt},clip]{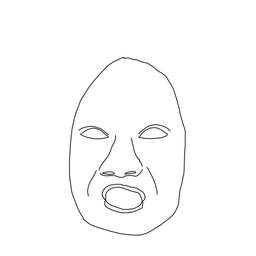}
        \put(0,-9){\includegraphics[width=0.06\linewidth,trim={150pt 200pt 150pt 100pt},clip]{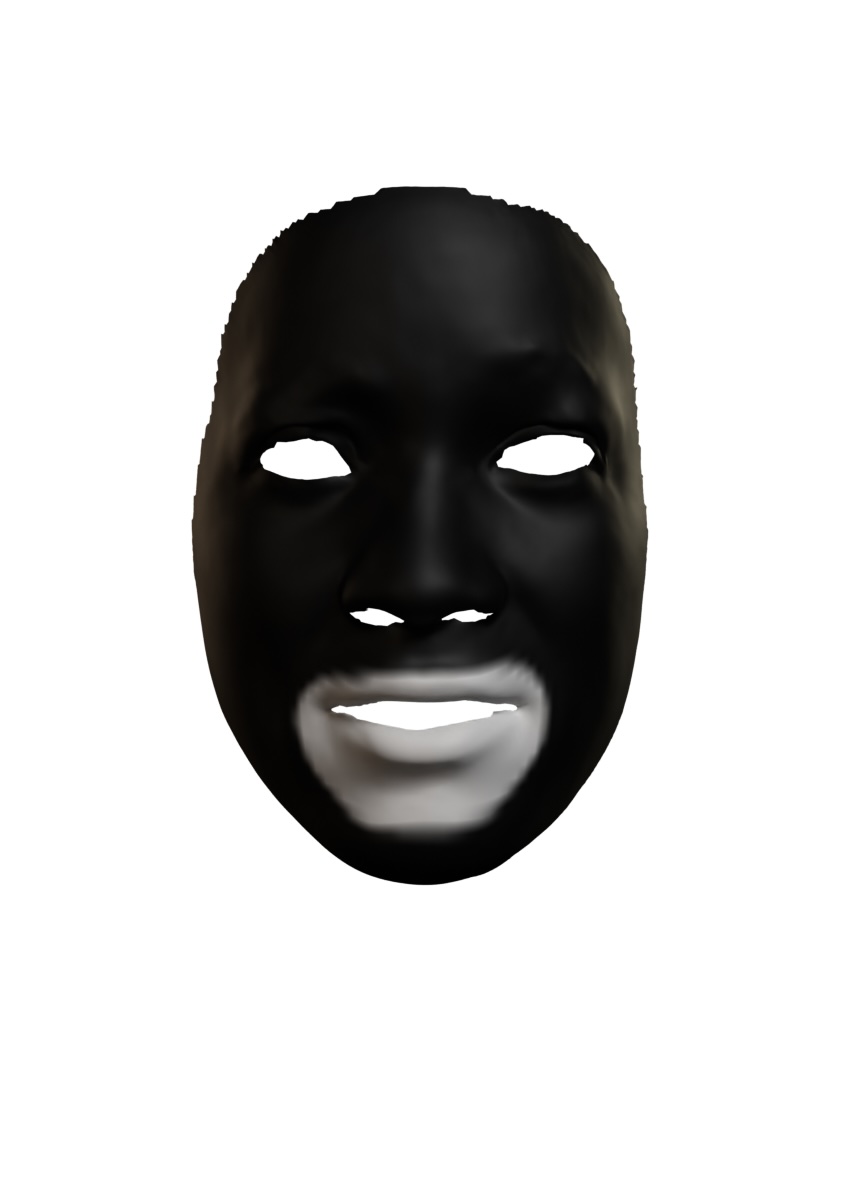}}
    \end{overpic}
     &
    \begin{overpic}[width=0.2\linewidth,trim={0 0 0 10pt},clip]{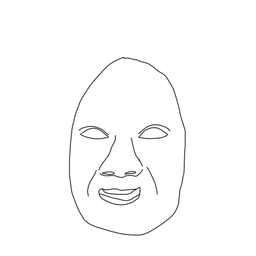}
        \put(0,-9){\includegraphics[width=0.06\linewidth,trim={150pt 200pt 150pt 100pt},clip]{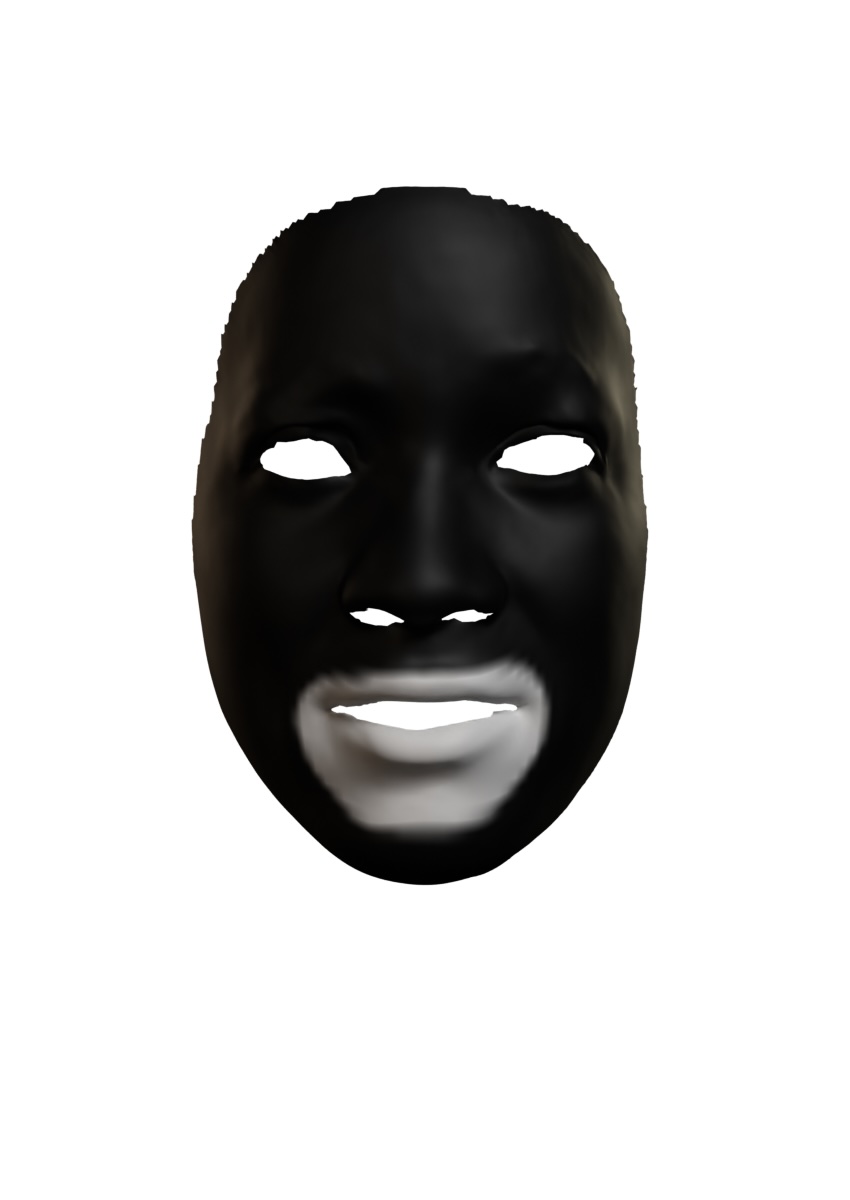}}
    \end{overpic}
     &
    \begin{overpic}[width=0.2\linewidth,trim={0 0 0 10pt},clip]{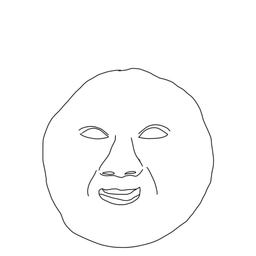}
        \put(0,-9){\includegraphics[width=0.06\linewidth,trim={150pt 200pt 150pt 100pt},clip]{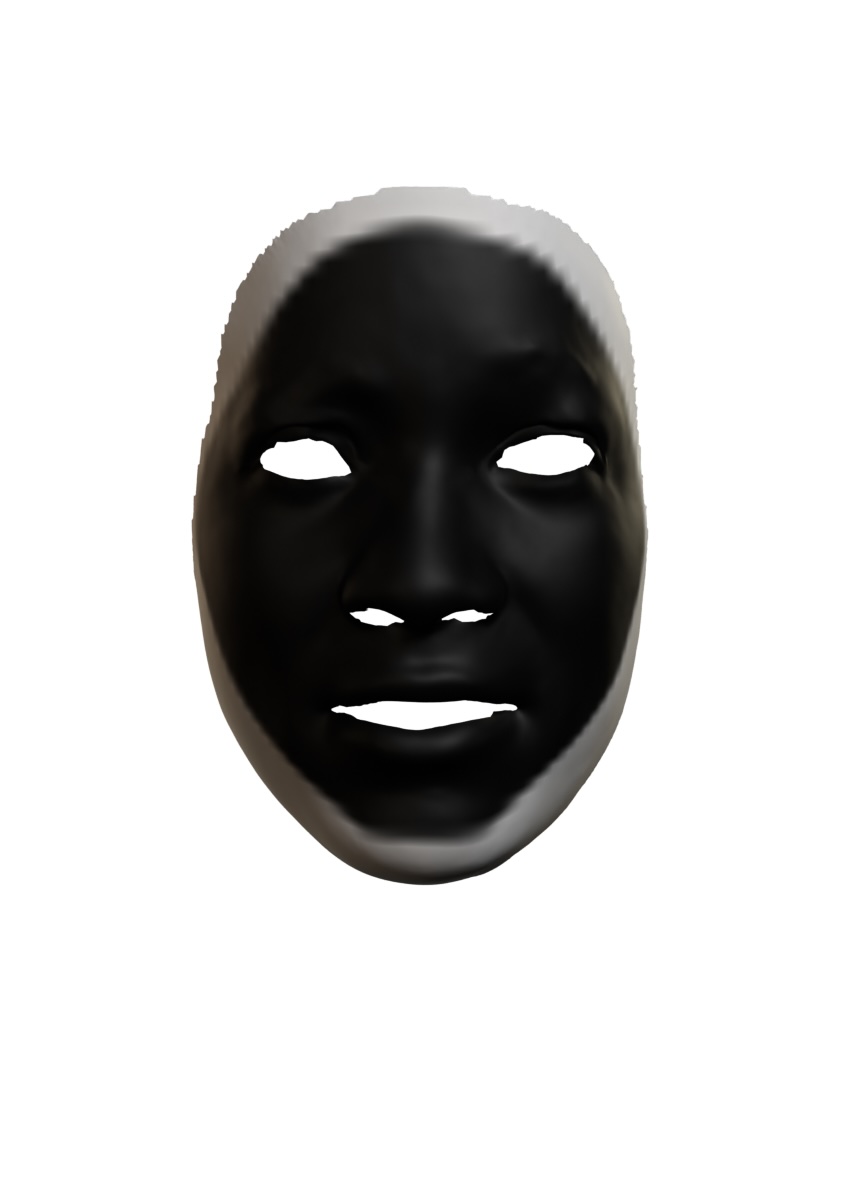}}
    \end{overpic}
     &
    \begin{overpic}[width=0.2\linewidth,trim={0 0 0 10pt},clip]{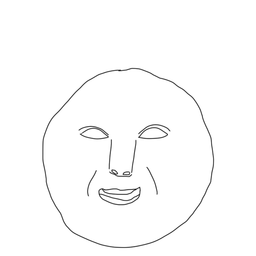}
        \put(0,-9){\includegraphics[width=0.06\linewidth,trim={150pt 200pt 150pt 100pt},clip]{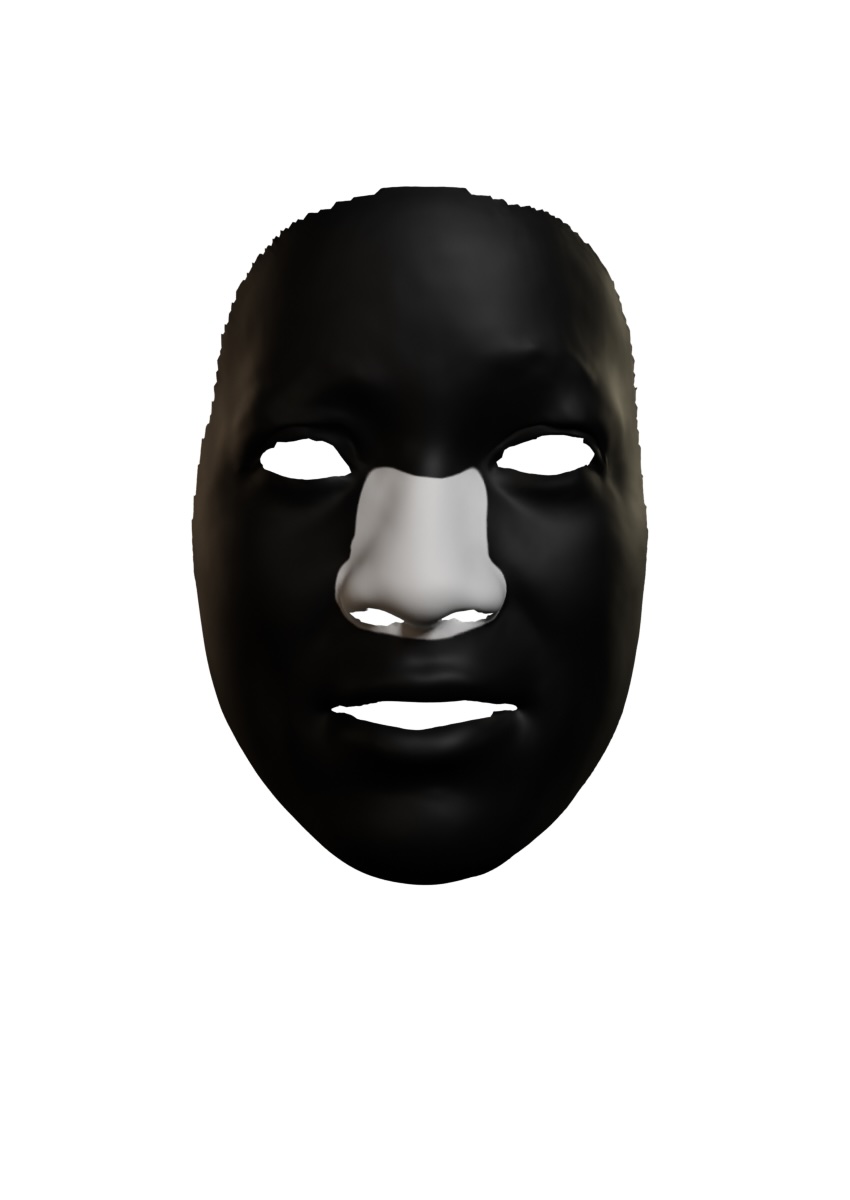}}
    \end{overpic}
    \\
    \begin{overpic}[width=0.2\linewidth,trim={0 100pt 0 10pt},clip]{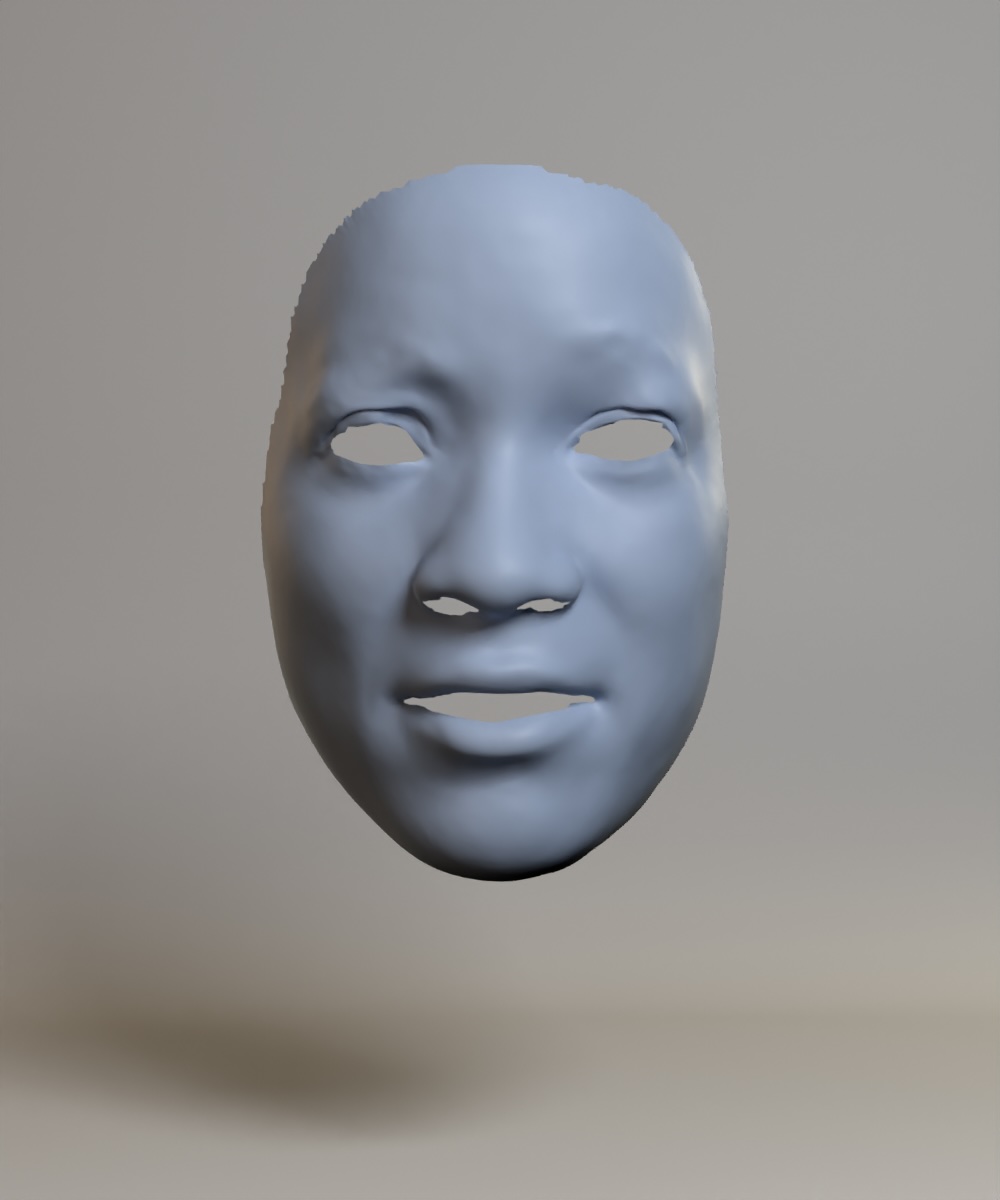}
        \put(2,5){a)}
    \end{overpic}
     &
    \begin{overpic}[width=0.2\linewidth,trim={0 100pt 0 10pt},clip]{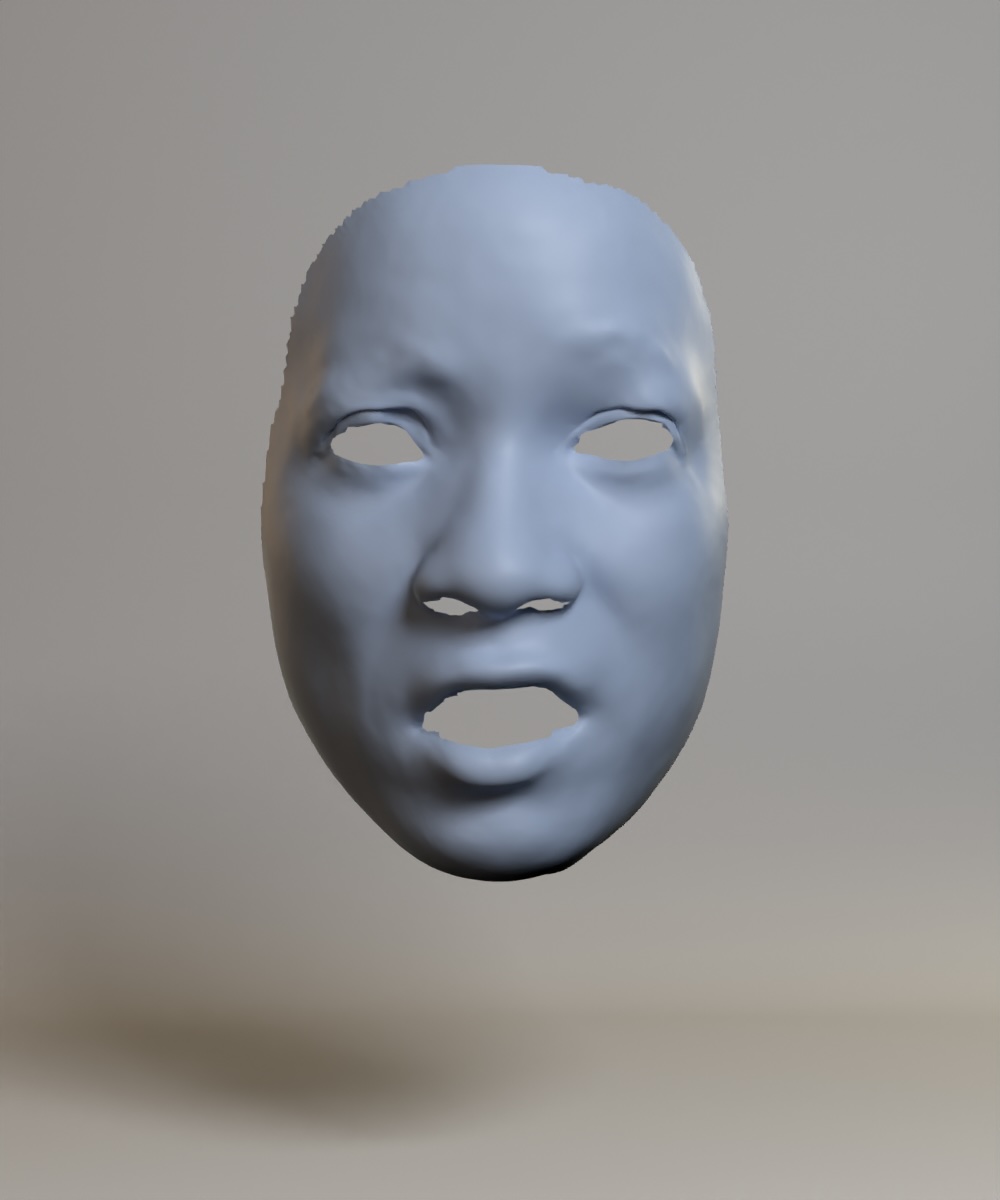}
        \put(2,5){b)}
    \end{overpic}
     &
    \begin{overpic}[width=0.2\linewidth,trim={0 100pt 0 10pt},clip]{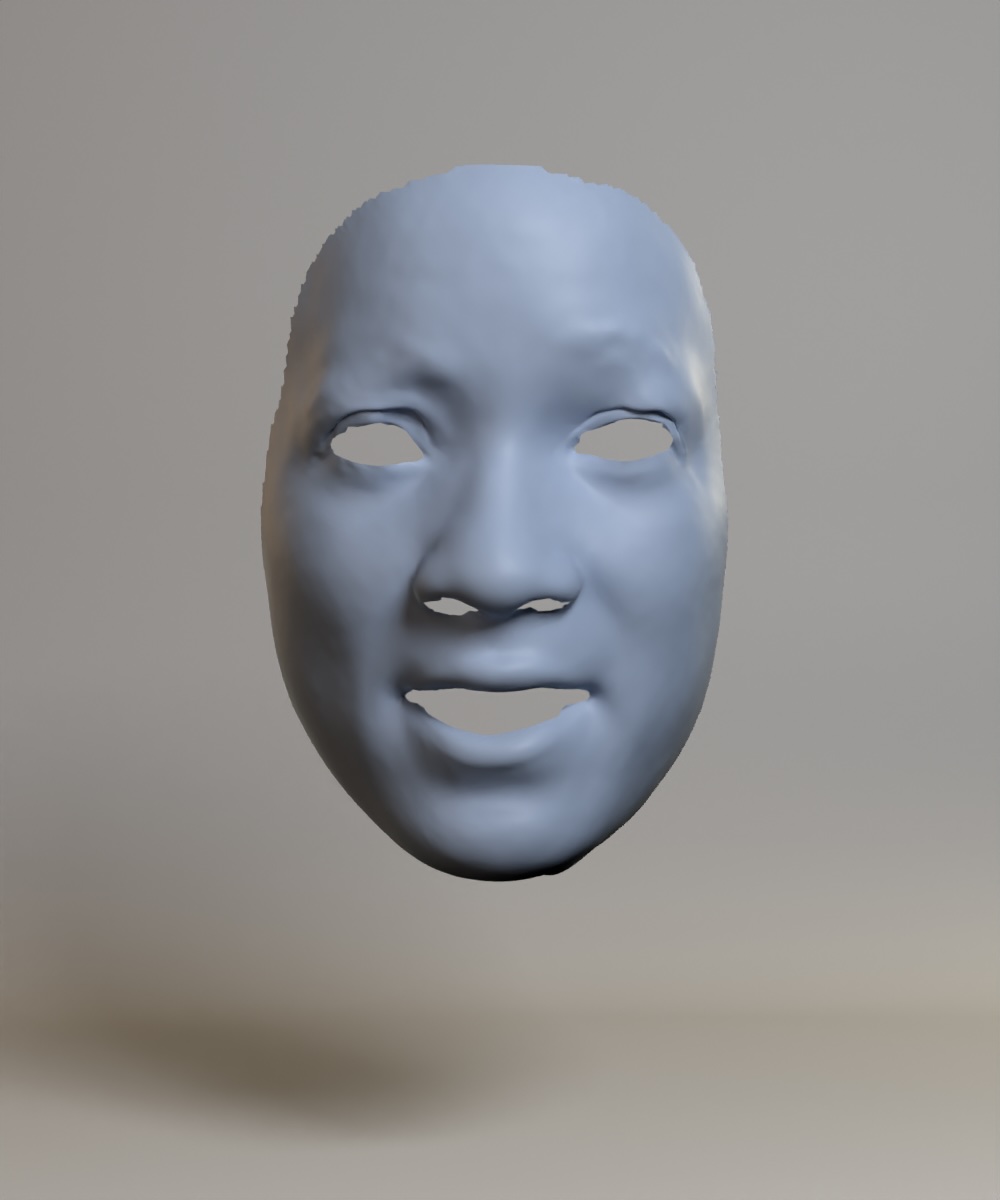}
        \put(2,5){c)}
    \end{overpic}
     &
    \begin{overpic}[width=0.2\linewidth,trim={0 100pt 0 10pt},clip]{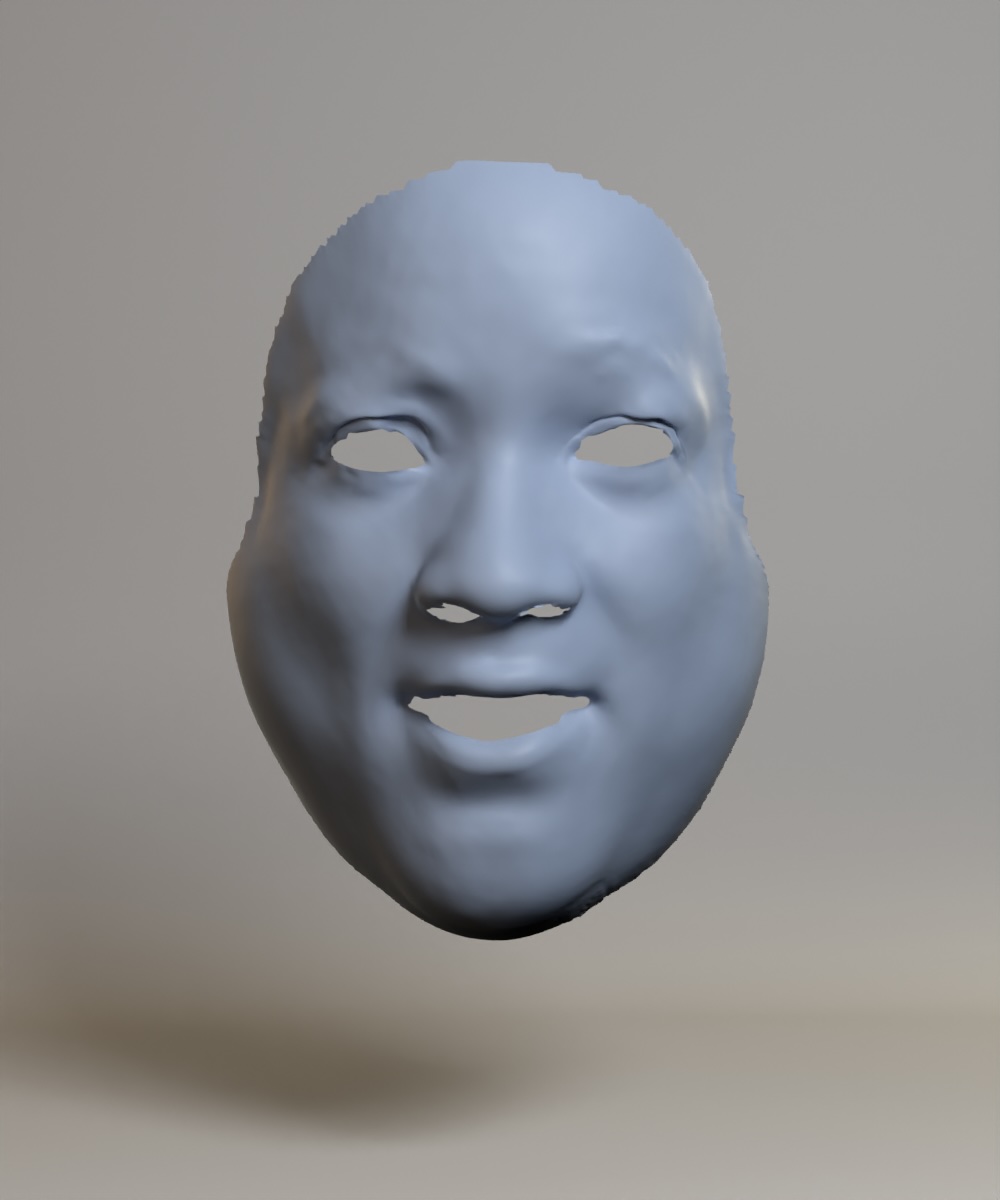}
        \put(2,5){d)}
    \end{overpic}
     &
    \begin{overpic}[width=0.2\linewidth,trim={0 100pt 0 10pt},clip]{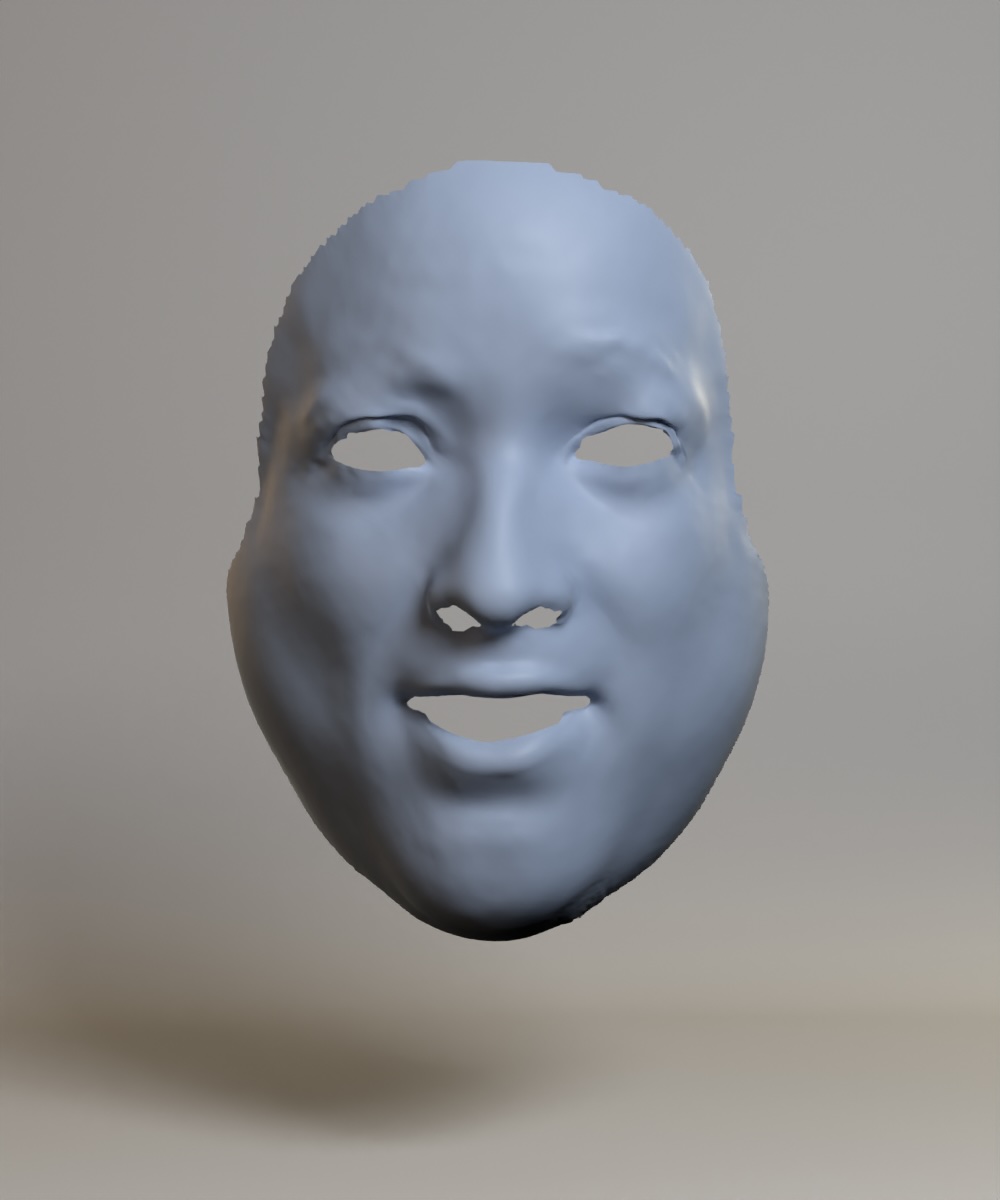}
        \put(2,5){e)}
    \end{overpic}
\end{tabular}
	\caption{Conditional generation from an input sketch (a), followed by local edits of the mouth (b, c), the face shape (d) and the nose (e). The masks used to constrain the region of modification are shown in the insets.}
	\label{fig:sketchedit}
\end{figure}

The latent space of our autoencoder preserves the spatial layout of the original UV position map, much like how the latent space of the image autoencoder in text-to-image models \cite{ldm} preserves the spatial layout of the encoded image. As a consequence, by masking regions in the latent UV position map corresponding to regions we wish to modify, and by denoising the masked regions, one can apply intuitive edits to particular regions of the facial geometry. Please refer to RePaint~\cite{repaint} for more details on the masking process. Even when using masks with sharp boundaries, the denoising process can take care of smoothly interpolating at the mask boundaries. We show results of guiding the editing of facial geometry with user conditions in ~\figref{fig:sketchedit}. Specifically, we show an interactive sketching workflow ($\sim$6 seconds/sample), where an artist can progressively edit a generated geometry by modifying one region at a time. 

\subsection{Dynamic Generation}
\label{subsec:dynamicgeneration}
\begin{figure}
	\centering
	\includegraphics[width=1.0\columnwidth]{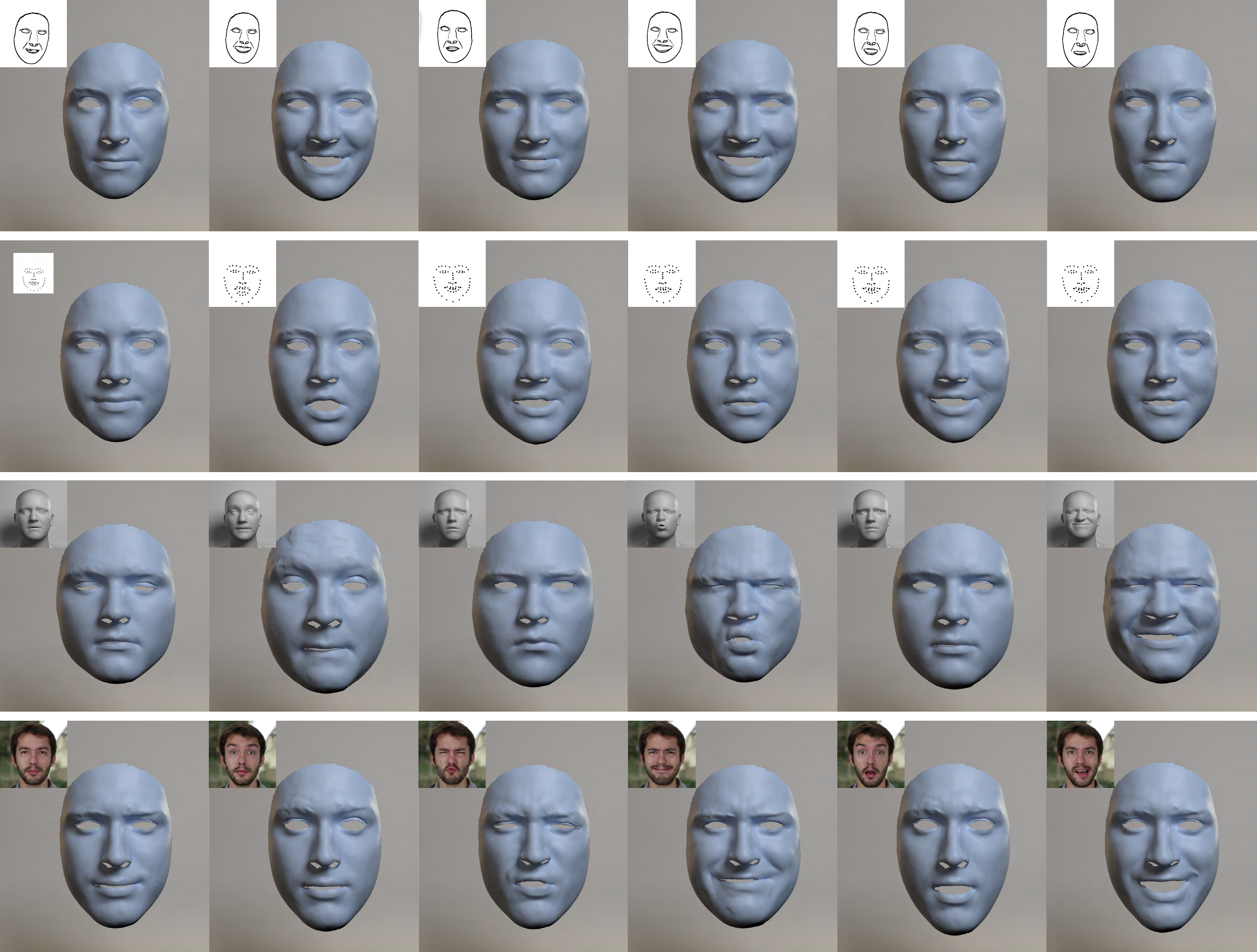}
	\caption{Dynamic geometry generation results given sketch, landmark, FLAME parameters or portrait photos from 4 different input videos as conditionings. Our results change smoothly across time while maintaining a consistent identity rather well.}
	\label{fig:dynamicgen}
\end{figure}

Although our model is only trained with static face shapes, we find that it can generate temporally stable 3D facial geometries when conditioned on per-frame FLAME parameters derived from animation sequences or on CLIP embeddings obtained from individual frames from in-the-wild videos. In \figref{fig:dynamicgen}, we show the generated 3D face geometry produced by our method when conditioned on various signals derived from videos.  To demonstrate the use of sketches as dynamic conditioning, we use a recent face reconstruction technique~\cite{Chandran2023} to track the facial geometry in 3D from an in-the-wild video and then render out 2D sketches using a hand-painted texture map. We identify that the only pre-processing required to obtain dynamically stable  generations from CLIP embeddings is to temporally smooth them before using them as the conditioning signal. To further ensure stable generations across time, we use the same noise seed and DDIM sampling.

\subsection{Limitations and Future Work}
\label{subsec:limitations}
As limitations, we identify that our model can produce geometric artifacts for extreme expressions, especially when controlled using FLAME's jaw pose parameters. This problem is mainly a data limitation and could be resolved by sourcing a larger dataset of extreme expressions, by oversampling expressions during training or by weighting the loss towards focusing more on extreme expressions. Additionally, we identify a limitation of extremely similar CLIP-based conditionings for left/right mirrors of asymmetrical face expressions, leading to a direction ambiguity in the geometry output. We refer to our supplementary material for further discussion and visualization of these failure cases. Beyond addressing those limitations, future work could incorporate facial appearance information into our method, enabling multimodal control over 3D faces with corresponding texture.

\section{Conclusion}
\label{sec:conclusion}
We propose a new framework for 3D facial geometry generation based on a latent diffusion model that can be guided using multiple types of conditionings (prompts). Our conditional geometry generator operates in a latent geometry space. It can produce high quality geometry at comparably fast inference speeds using a UV position map representation. It can be seamlessly conditioned on hand-drawn sketches, 2D landmarks, Canny edges, FLAME-parameters, RGB portrait photos and text; resulting in a comprehensive facial geometry generator that supports many applications. For example, stochastic detail variation in the generated geometry or local geometry edits. We train our model from scratch on only static face shapes captured in a studio setting and yet demonstrate that our model can generalize reasonably to in-the-wild conditioning signals, and can also generate facial performances when conditioned on frames from video data.

\bibliographystyle{cag-num-names}
\bibliography{diffusion-faces}

\clearpage

\appendix
\renewcommand\thesection{\Alph{section}} 
\counterwithout{figure}{section}
\counterwithout{table}{section}

This supplementary document and our supplementary video provide additional insights into our method and results.

\section{Ablation Studies}
\label{sec:ablation}

\begin{figure}[h]
	\includegraphics[width=1.0\columnwidth]{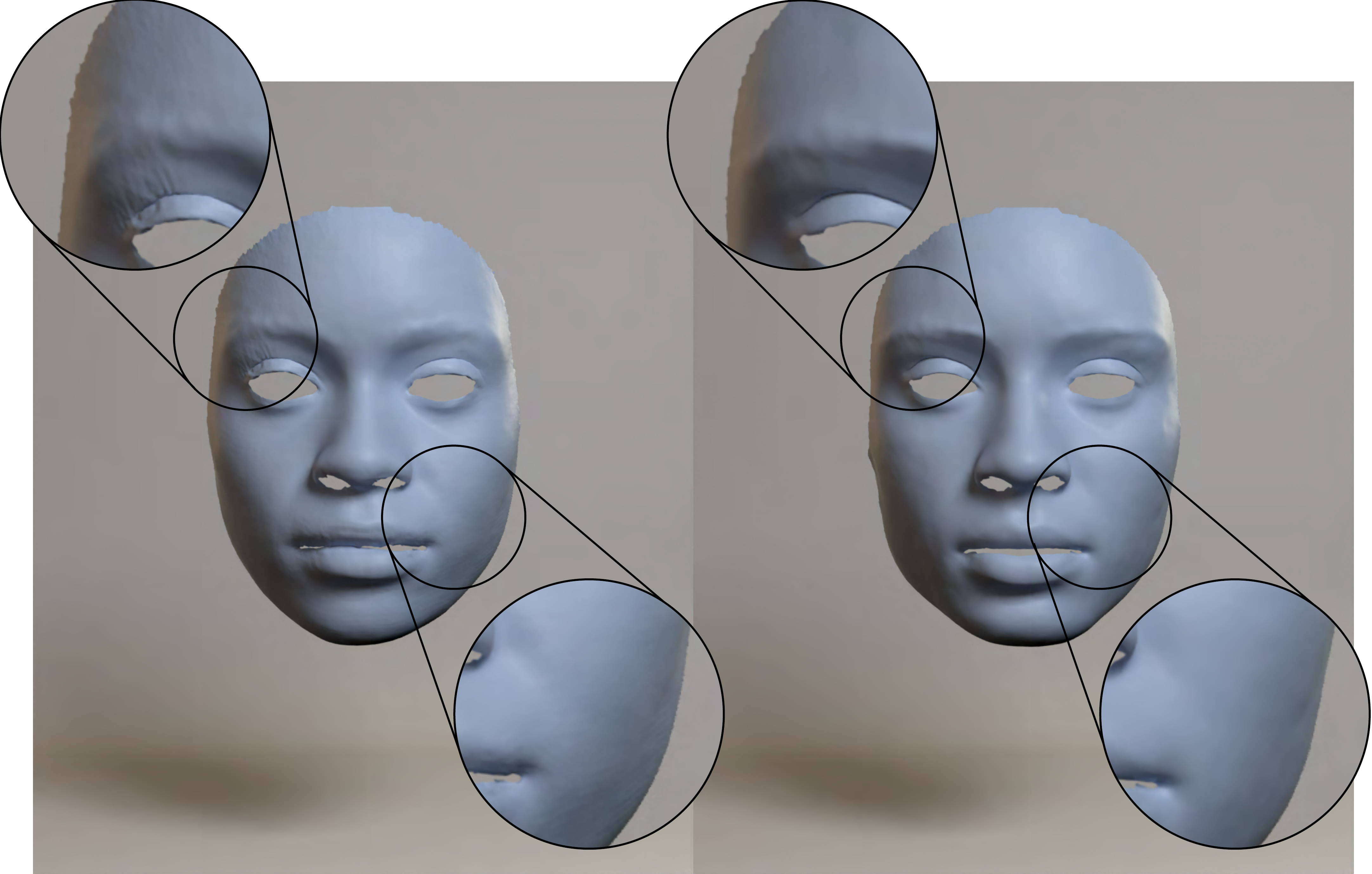}
	\caption{Ablation study of the geometry representation. On the left is the generated geometry when training on the full vertex map representation, which shows visible artifacts (e.g., on the eyelid). Our proposed training with the delta vertex maps (right) removes such artifacts.}
	\label{fig:vertexmap}
\end{figure}

To validate the benefits of representing the face geometry as a delta from the template mesh, we train a latent diffusion model (LDM) on the full vertex map representation and a second LDM on the delta vertex map representation. We qualitatively compare both representations in \figref{fig:vertexmap} and can observe that the delta vertex map representation leads to less artifacts compared to the full vertex map representation (\eg on the eyelids). \\

Next, we validate the benefits of adding geometry data augmentations to our training data.  We compare FLAME parameter conditioned generations from two LDMs, where one was trained with and the other was trained without geometry data augmentations. As we illustrate in \tabref{tab:augmentation_ablation}, our generations are closer to the ground truth geometry (lower vertex-to-vertex error), when using geometry data augmentations. This result indicates that using geometry data augmentations improves the models ability to capture unseen identities.

\begin{table}[h]
	\centering
	\caption{\label{tab:augmentation_ablation} We compare the 3D face geometry generated by diffusion models that were trained with and without 3D data augmentations. We measure the vertex-to-vertex error (V2V) in mm between FLAME parameter conditioned generations and ground truth geometry on neutral shapes from our validation set. The model trained using data augmentation is able to capture unseen identities better. Results are averaged over three different seeds. } 
	\begin{tabular}{l c c c}
		\toprule
		\textbf{V2V error} & \textbf{Mean $\downarrow$} & \textbf{Median $\downarrow$} & \textbf{Std $\downarrow$}  \\
		\midrule
		No augmentations & 4.093 & 3.692 & 2.170 \\
		With augmentations
		& \textbf{3.757} & \textbf{3.352} & \textbf{2.085} \\
		\midrule
	\end{tabular}
\end{table}

\section{Implementation Details}
\label{sec:implementation_details}
\begin{figure}[h]
	\centering
	\includegraphics[width=0.7\columnwidth]{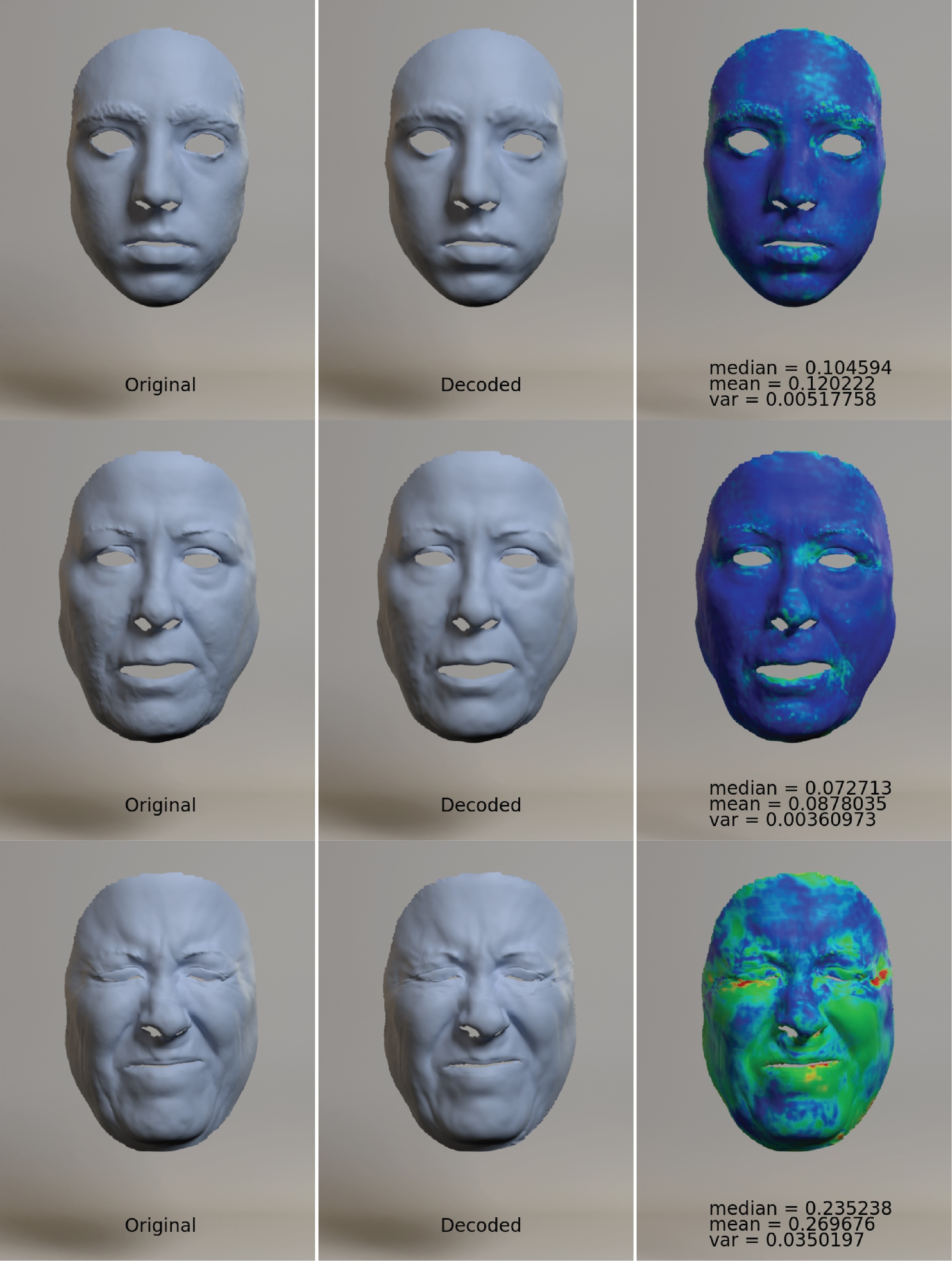}
	\caption{The VAE reconstruction error is usually below 0.3 millimeters when compared to the ground truth geometry. Some high frequency details are lost after encoding and decoding the original geometry with the VAE due to VAE compression.}
	\label{fig:ae-reconstruction}
\end{figure}

\begin{figure}[h]
	\centering
	\includegraphics[width=1.0\columnwidth]{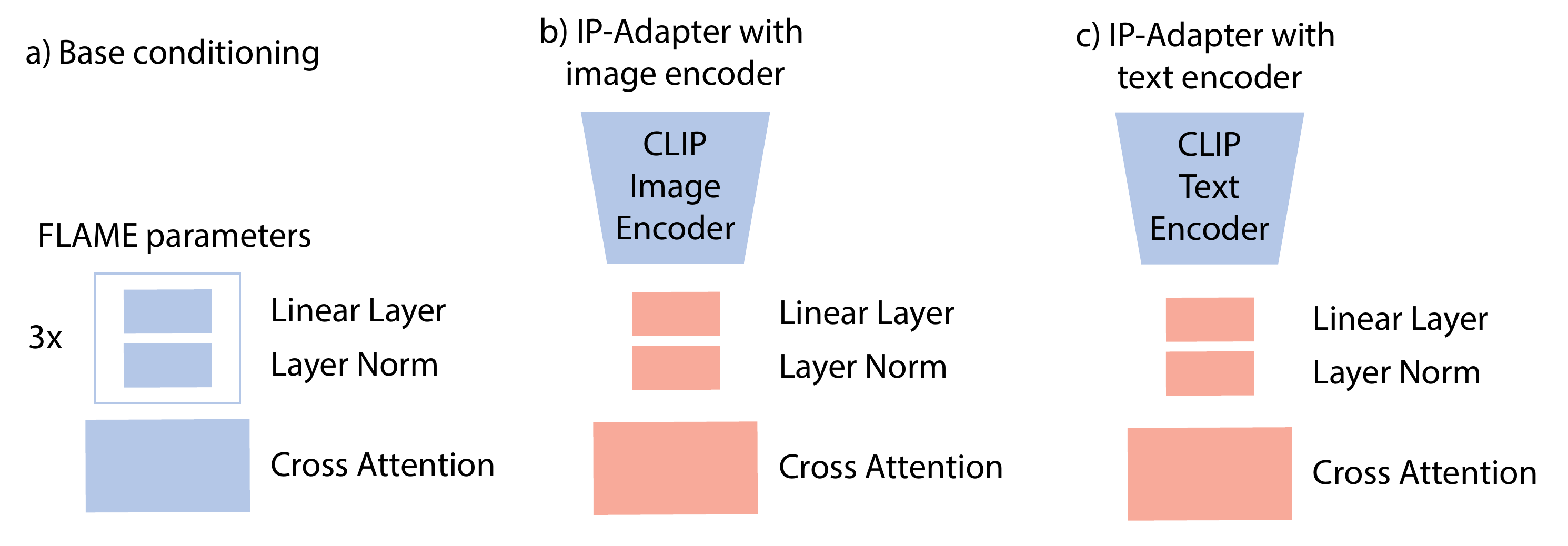}
	\caption{Modality injection visualization. Our FLAME parameter base conditioning layers (a) are trained jointly with the diffusion model. Afterwards, both are frozen and only the new mode-specific layers are trained. Sketches, portrait photos, Canny edges and landmarks are passed through a frozen CLIP image encoder (b) before reaching their own set of mode-specific trainable layers (one set of layers per mode). Text is passed through a frozen CLIP text encoder (c) before reaching a set of text-specific trainable layers.}
	\label{fig:cross-attention}
\end{figure}

For our dataset, we crop the full head face geometry to allocate more vertices to the face region, representing 50520 face vertices within each $256^{2}$ UV position map. At $256^{2}$ resolution we can represent reasonably high-resolution face geometry while being able to limit our VAE training time to 8 days using our training dataset of 7752 samples ($\sim$1.4 seconds/iteration; batch size 8). We use a learning rate of 4.5e-6 and a codebook size of 8192. Note that for training our VAE, we did not use the data augmentations described in \secref{sec:ablation} as the autoencoder was already able to reconstruct test geometries with high accuracy when trained only on the studio dataset. The vertex error between reconstructions and the original geometry is usually below 0.3 millimeters and only very high-frequency details are lost. We visualize the VAE reconstruction error in \figref{fig:ae-reconstruction}. Next, we train our LDM for 4 days ($\sim$1.6 seconds/iteration; batch size 12) with a learning rate of 1e-4 and diffusion timesteps $T=1000$. We utilize geometry data augmentations with corresponding FLAME fits during training (+200k samples) to allow for better generalization across identities during generation. Afterwards, we train each set of cross-attention layers with a learning rate of 1e-4 for 6 days ($\sim$3.3 seconds/iteration; batch size 24) while keeping the LDM frozen. With a probability of 0.05, we randomly set either $\mathbf{c}_0$ or $\mathbf{c}_m$ or both to their null embeddings during training. This step enables classifier-free guidance at inference. Note that we do not add augmented geometry data to train the new cross-attention layers because we do not have access to paired mode-geometry data for modes such as portrait photos. We use the CLIP ViT-L/14 model~\cite{clip} to extract 768 dimensional CLIP feature vectors as our conditioning representation for all modalities except the base FLAME parameter conditioning. We visualize the layers that pass our base FLAME parameter conditioning to the diffusion model in \figref{fig:cross-attention}. These layers are trained jointly with the diffusion model parameters. Afterwards, the diffusion model and the base conditioning layers are frozen. For adding a new modality a newly added set of mode-specific layers are trained (linear layer, layer norm and a set of cross-attention layers). Sketches, portrait photos, Canny edges and landmarks are passed through a frozen CLIP image encoder before reaching their own mode-specific trainable layers. We train a different set of layers per mode (e.g. one for sketches, one for portrait photos etc.). Text is passed through a frozen CLIP text encoder before reaching text-specific trainable layers (\figref{fig:cross-attention}). All training experiments were run on a single RTX A6000 GPU. Inference was run on single RTX A6000 GPUs, 3090 GPUs, and 1080 GPUs. Also note that the VAE and the LDM have to be trained only once and novel conditioning modes can be added by training only the new cross-attention layers. Unless mentioned otherwise, we generate every result by running DDPM sampling steps $S=50$ with conditioning strength $w=1$. The average time to generate a geometry sample with our diffusion model is $\sim$6 seconds on a single 3090 GPU. To aid the visual similarity for the comparison with the state-of-the-art methods, we complete the head by deforming a template head to match our generated face. We run the CLIP score evaluation for all methods on full head renders. We align all 3D faces to the same space, before rendering each with the same camera. We use the CLIP ViT-B/32 variant for the score calculation and report the average score for each method.

\section{Exact Text Prompts}

Table~\ref{tab:text_prompts} lists the exact text prompts used for the CLIP score comparison and the figures in the main document. 

\begin{table*}
	\caption{\label{tab:text_prompts} The exact text prompts used in the comparison to the related work methods. Prompts 1 - 10 specify a neutral expression, while prompts 11 - 20 specify other facial expressions.} 
	\centering
	\begin{tabular*}{\textwidth}{l p{0.8\textwidth}}
		\toprule
		\textbf{Nr.} & \textbf{Text prompt}\\
		\midrule
		1
		& A shaded, textureless 3D face model of an African woman with a neutral expression.\\
		\midrule
		2 
		& A shaded, textureless 3D face model of an overweight man with a neutral expression.\\
		\midrule
		3 
		& A shaded, textureless 3D face model of an old woman with a neutral expression.\\
		\midrule
		4
		& A shaded, textureless 3D face model of a middle-aged Asian person with a neutral expression.\\
		\midrule
		5 
		& A shaded, textureless 3D face model of a middle-aged Caucasian woman with a neutral expression.\\
		\midrule
		6 
		& A shaded, textureless 3D face model of a woman with high cheekbones, a defined jawline, and a straight nose with a neutral expression.\\
		\midrule
		7 
		& A shaded, textureless 3D face model of a young woman with a round face, big eyes and small mouth with a neutral expression.\\
		\midrule
		8 
		& A shaded, textureless 3D face model of a man with a chubby face, a wide forehead and a wide nose with a neutral expression.\\
		\midrule
		9  
		& A shaded, textureless 3D face model of a young African man with a neutral expression.\\
		\midrule
		10 
		& A shaded, textureless 3D face model of a young Asian man with a neutral expression.\\
		\midrule
		\midrule
		11 
		& A shaded, textureless 3D face model of a smiling overweight man.\\
		\midrule
		12 
		& A shaded, textureless 3D face model of an overweight man shouting angrily.\\
		\midrule
		13 
		& A shaded, textureless 3D face model of a sad Caucasian man.\\
		\midrule
		14 
		& A shaded, textureless 3D face model of a middle-aged Asian person with a kiss face expression.\\
		\midrule
		15  
		& A shaded, textureless 3D face model of a smiling African woman.\\
		\midrule
		16
		& A shaded, textureless 3D face model of a woman with high cheeckbones, a defined jawline, and a straight nose. Her mouth is opened to the side.\\
		\midrule
		17 
		& A shaded, textureless 3D face model of an angry Caucasian man.\\
		\midrule
		18 
		& A shaded, textureless 3D face model of a man with a chubby face, a wide forehead and a wide nose with a big smile on his face.\\
		\midrule
		19 
		& A shaded, textureless 3D face model of a young African man with a closed eyes facial expression.\\
		\midrule
		20
		& A shaded, textureless 3D face model of a young Asian man with a very surprised facial expression. His eyes and mouth are wide open and the eyebrows raised.\\
		\midrule
		\midrule
	\end{tabular*}
\end{table*}

\section{Additional Geometry Editing Results}

\begin{figure}
	\includegraphics[width=1.0\columnwidth]{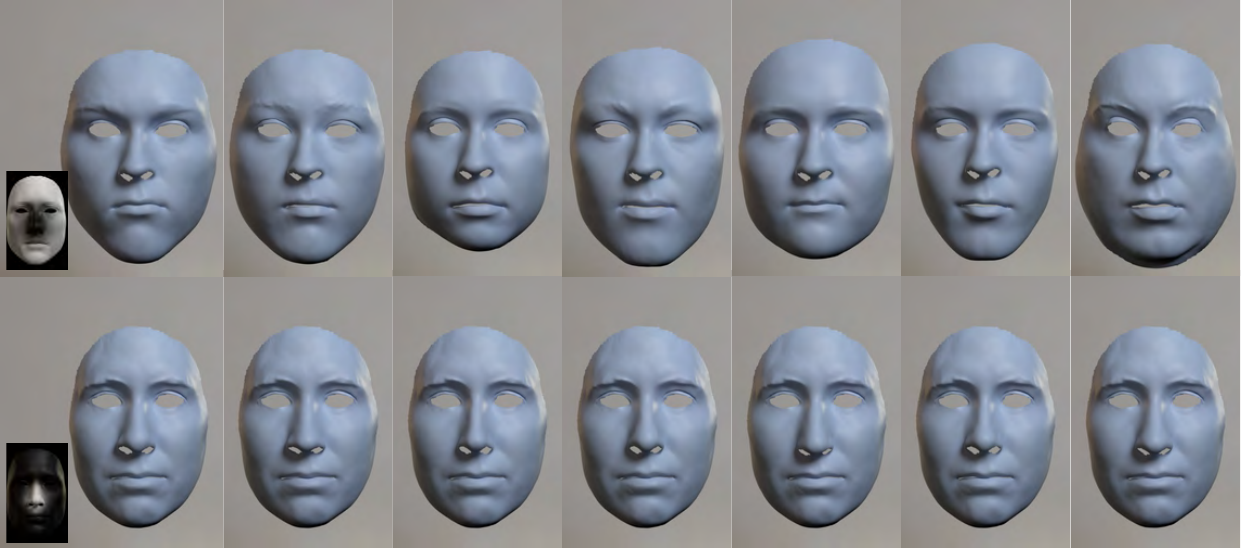}
	\caption{Unconditional face shape editing (inpainting). In the top row, the nose is kept fixed, while we sample the remaining regions unconditionally. In the bottom row, we sample the nose region unconditionally while keeping the other regions fixed.}
	\label{fig:inpainting}
\end{figure}

Further mask-based editing of facial geometries using our model is shown in ~\figref{fig:inpainting}. In the top row of ~\figref{fig:inpainting} we mask the nose region of the latent position map, such that it remains fixed throughout the multiple steps of denoising. We then generate multiple geometry samples by varying the initial noise input to the diffusion model. The noise predicted at each denoising step is multiplied with the nose mask before being fed as input to the denoising UNet for the next time step. This denoising procedure leads to generations where the generated samples all share the same nose shape, but vastly differing facial identities. In the bottom row of ~\figref{fig:inpainting}, we show the result of inverse masking, where the face shape is held fixed while allowing the nose shape to change. Our model produces meaningful results in both cases. 

\section{Additional Quantitative and Qualitative Results}

\begin{figure}
	\includegraphics[width=1.0\columnwidth]{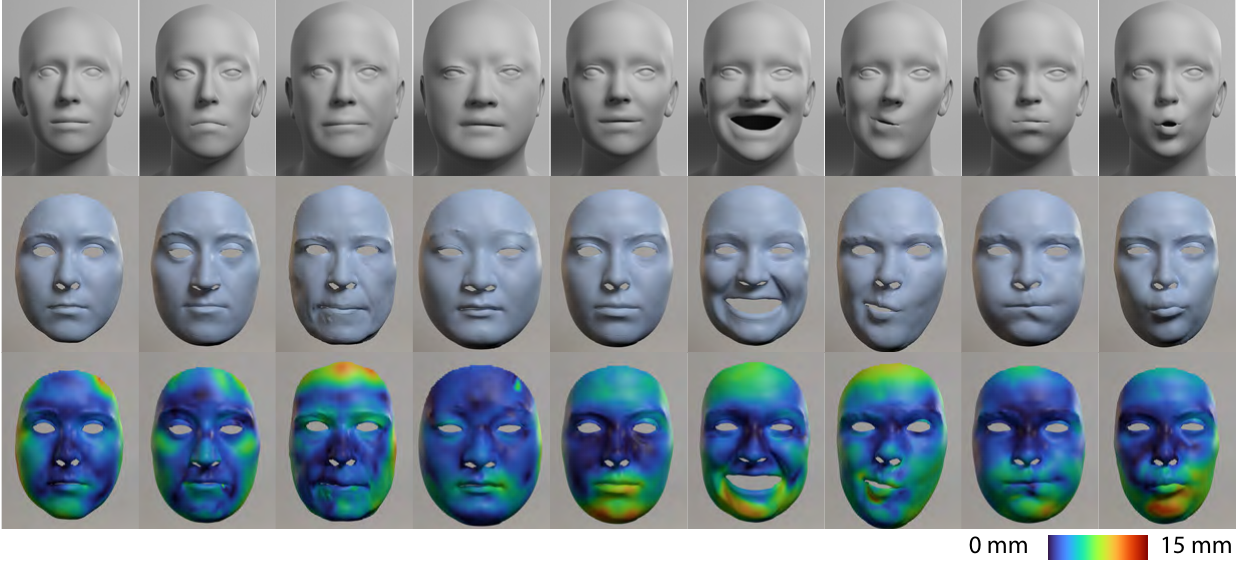}
	\caption{Error maps on our validation set. The first row shows the FLAME mesh as generated by the FLAME face model from the input FLAME parameters. The second row shows the generated geometry from our model conditioned on the respective FLAME parameters. The third row visualizes the error from our conditional generations to the original scanned geometry in our validation set. The first four columns are various identities, while the last five columns are different expressions of the same subject.}
	\label{fig:validationerror}
\end{figure}

For the base FLAME parameter conditioning, we visualize the error maps to the ground truth scanned geometry in \figref{fig:validationerror}.
Next, we compare the text-to-geometry generation results of HeadArtist~\cite{headartist} and HumanNorm~\cite{humannorm} with our method. Both are based on Deep Marching Tetrahedra~\cite{dmtet} and SDS optimization~\cite{sdsloss} and can represent face parts beyond the skin. The extracted face geometries differ in topology and optimizing for one sample takes around one hour on a 3090 GPU. In contrast, our method's inference speed is 1000-times faster on a 3090 GPU and produces results in a single common topology.

\section{Failure Cases}

\begin{figure}
	\centering
	\includegraphics[width=1.0\columnwidth]{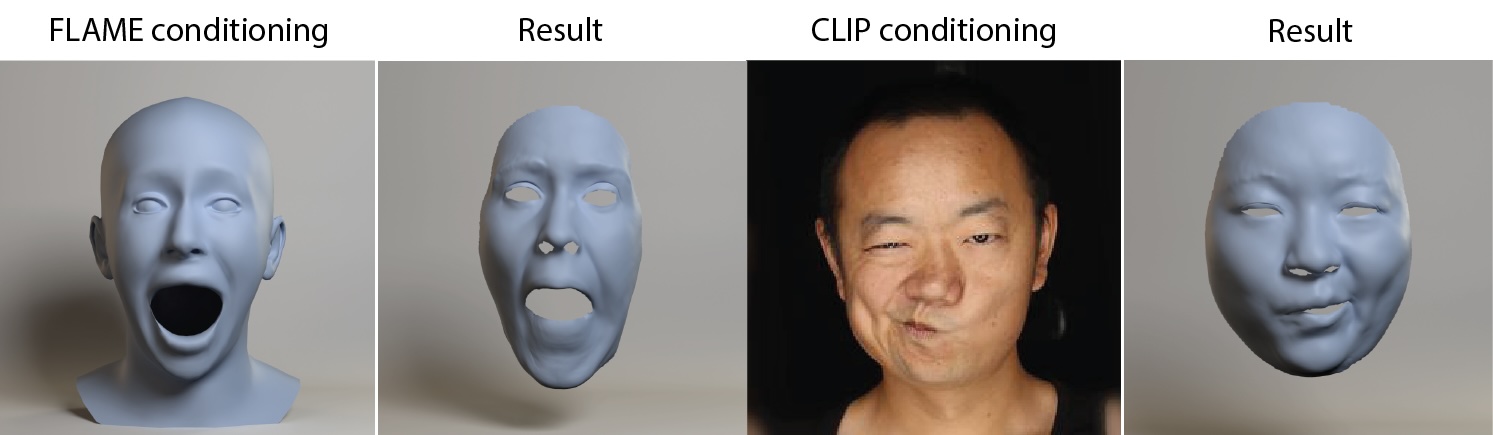}
	\caption{Failure cases. Our results can display geometric artifacts around the mouth area for extreme expressions (column 2) and sometimes incorrect mouth opening sides when conditioned on CLIP embeddings (column 4).}
	\label{fig:failure_cases}
\end{figure}

We do observe geometric artifacts around the mouth region for extreme expressions (Figure~\ref{fig:failure_cases}, column 2), due to limited extreme expressions in our training data. Additionally, the generated face geometry can open the mouth to the wrong side when conditioning with CLIP embeddings (Figure~\ref{fig:failure_cases}, column 3 and 4). We identify that this behavior occurs, when the CLIP embeddings for both mouth opening directions (left/right) are extremely similar (cosine similarity close to 1). Thus, this behavior is caused by the similarity of specific conditionings and not by the diffusion model. 

\begin{figure}
	\centering
	\includegraphics[width=.9\columnwidth]{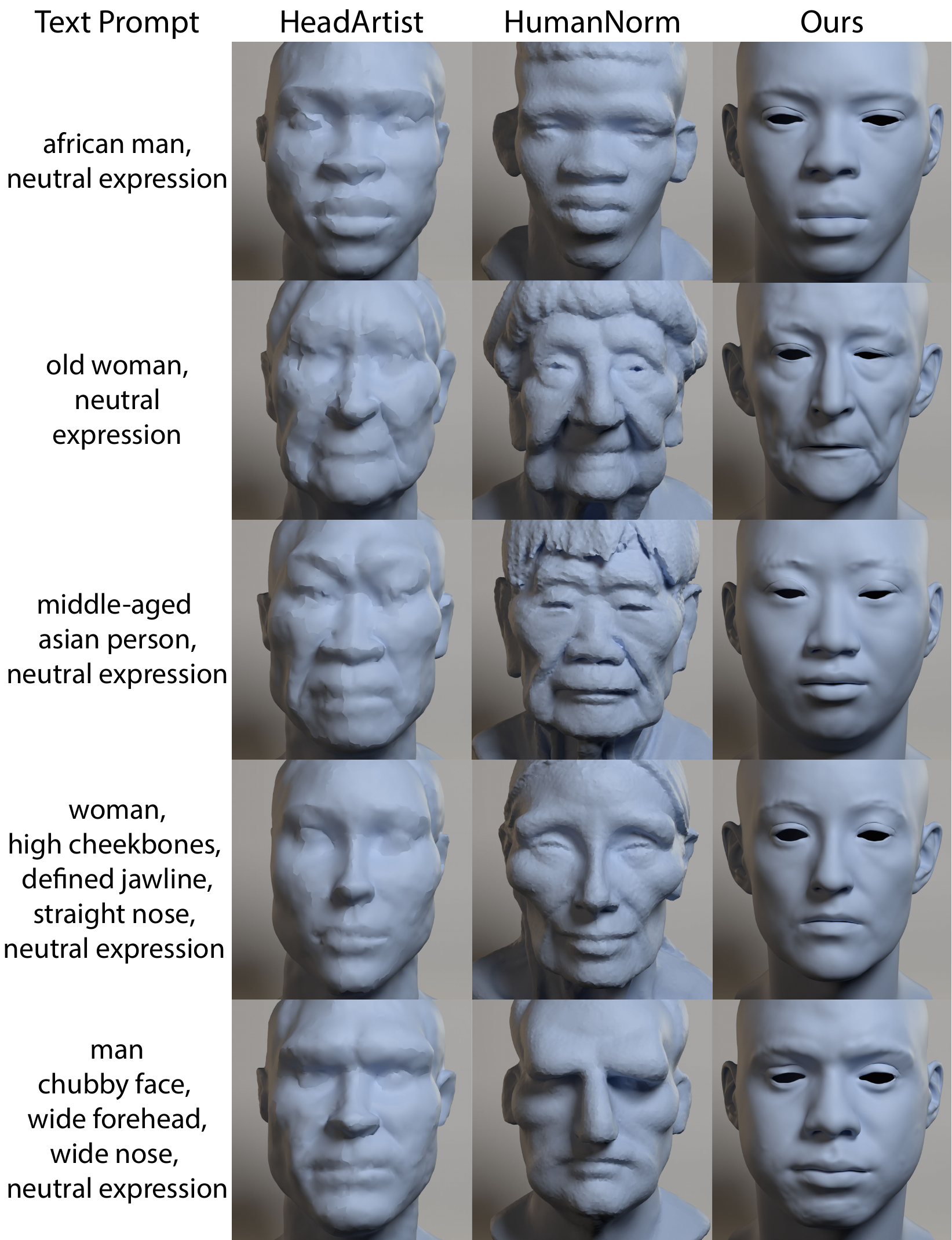}
	\caption{Qualitative comparison with the text-to-geometry generation ability of HeadArtist~\cite{headartist} and HumanNorm~\cite{humannorm}. For legibility, we shortened the text prompt. Please refer to Table~\ref{tab:text_prompts} for the exact text prompts.}
	\label{fig:dmtet_methods}
\end{figure}

\end{document}